\documentclass{article}
\usepackage{biblatex}
\usepackage{float}

\usepackage{graphicx,subcaption}

\nocite{*}

\usepackage[english]{babel}

\usepackage[letterpaper,top=2cm,bottom=2cm,left=3cm,right=3cm,marginparwidth=1.75cm]{geometry}

\usepackage{xcolor}
\usepackage{listings}

\lstdefinestyle{DOS}
{
    backgroundcolor=\color{black},
    basicstyle=\scriptsize\color{white}\ttfamily
}

\usepackage{authblk}
\usepackage{blindtext}
\usepackage{amsmath}
\usepackage{graphicx}
\usepackage[colorlinks=true, allcolors=blue]{hyperref}

\title{AutoVisual Fusion Suite: A Comprehensive Evaluation of Image Segmentation and Voice Conversion Tools on HuggingFace Platform}
\author{\parbox{.5\textwidth}{\centering
  Amirreza Hashemi \\
  \small Email: \href{mailto:amirrezahmi2002@gmail.com}{amirrezahmi2002@gmail.com} \\[3pt]
  Github: \href{https://github.com/Amirrezahmi}{Amirrezahmi}\\
  }}
\affil{Undergraduate Student in Computer Science, Central Tehran Branch, Islamic Azad University, Tehran, Iran}
\begin{document}
\maketitle

\begin{abstract}
This study presents a comprehensive evaluation of tools available on the HuggingFace platform for two pivotal applications in artificial intelligence: image segmentation and voice conversion. The primary objective was to identify the top three tools within each category and subsequently install and configure these tools on Linux systems. We leveraged the power of pre-trained segmentation models such as SAM and DETR Model with ResNet-50 backbone for image segmentation, and the so-vits-svc-fork model for voice conversion. This paper delves into the methodologies and challenges encountered during the implementation process, and showcases the successful combination of video segmentation and voice conversion in a unified project named AutoVisual Fusion Suite.
\end{abstract}

\section{Introduction}

This research presents a meticulous exploration of the tools available on the HuggingFace platform, focusing on two critical applications, image segmentation and voice conversion. These tasks demand the application of sophisticated AI algorithms and techniques to achieve the desired outcome. Image segmentation involves dividing an image into multiple segments or sets of pixels, often to simplify or change the representation of an image into something more meaningful and easier to analyze. On the other hand, voice conversion is a more nuanced task that involves transforming the characteristics of a source voice to mimic a target voice, while preserving the linguistic content.

The first phase of our study involved evaluating the existing toolchains on HuggingFace for Image Segmentation and Voice Conversion, with the goal of discerning the top three toolchains within each category. Following this, we endeavored to aid engineers in setting up the Python toolchain on Linux for the selected toolchains.

Our research culminated in the creation of the AutoVisual Fusion Suite, an open-source project that combines video segmentation and voice conversion into a unified solution. The results of this research will serve as a valuable resource for professionals and enthusiasts in the field of artificial intelligence, providing insights into the selection and utilization of tools for these crucial application categories.

\section{Video Segmentation chosen models}

In our task of video segmentation, we were unable to locate an open-source model on Hugging Face. This led us to explore image segmentation instead. Our objective in video segmentation was to input two videos – the first featuring individuals and the second showcasing a background. The target video would then incorporate individuals from the first video into the background of the second.

Unfortunately, there are limited published models for image segmentation on Hugging Face. This led us to consider working with images, otherwise known as image segmentation.

In image segmentation, we provide two images as input. For example, the first image could be a person and the second a background. Our target is to produce an image that places the person from the first image onto the background of the second. This task involves dividing an image into multiple segments or sets of pixels, often to simplify or change the representation of an image into something more meaningful and easier to analyze.

This approach allows us to provide different frames of a video as input, and the model can perform the necessary transformations. However, the accuracy of this method cannot be guaranteed as it depends on various factors.

If we wanted to input two videos into the model, we would separate the videos into frames. For example, if we have two videos each 10 seconds long with 60 frames per second, we would have 600 frames for the first video and 600 for the second. We would then input these pair of frames one by one into the model and collect the output.

The output frames can then be concatenated to create a single output, resulting in 600 frames in total. This approach could potentially reduce the model size and make it lighter, allowing for more flexibility in its use. However, this method does not consider the temporal relationship between frames.

\subsection{Image Segmentation}
Image Segmentation is a process that partitions an image into distinct segments, associating each pixel with a specific object. This task encompasses several variants, including instance segmentation, panoptic segmentation, and semantic segmentation. The Hugging Face website offers an array of models and techniques dedicated to image segmentation. Below, you'll find a selection of the models and techniques utilized:

\begin{enumerate}
\item Semantic Segmentation: Semantic segmentation is the task of segmenting parts of an image that belong to the same class. In this technique, models make predictions for each pixel and return the probabilities of the classes for each pixel. The models are evaluated using Mean Intersection Over Union (Mean IoU) \cite{hg}.
\item Instance Segmentation: Instance segmentation is a variant of image segmentation where every distinct object is segmented, instead of one segment per class \cite{hg}.
\item Panoptic Segmentation: Panoptic segmentation is an image segmentation task that segments the image both by instance and by class, assigning each pixel a different instance of the class \cite{hg}.
\end{enumerate}

\subsection{Algorithm}

Our algorithm is as follows:
\begin{enumerate}
    \item Use the image segmentation model to convert the background of the target(s) to a transparent background.
    \item Add the converted image in step 1 to the second image.
    \item Repeat this process for each pair of video frames to create the target video.
\end{enumerate}

Our algorithm is designed to create a target video by applying image segmentation techniques. First, we utilize a model for image segmentation, such as the DEtection TRansformer (DETR), BEiT, SegFormer, etc., to extract the target(s) and convert its background to a transparent background. This step helps isolate the foreground object from the rest of the image. Once we have the transparent background image, we proceed to the next step.

Next, we add the transparent background image to a second image. This process involves overlaying the two images, with the transparent background image being placed on top of the second image. By doing so, we combine the foreground object from the first image with the background of the second image. This step helps create a composite image that includes the desired foreground object and the background from the second image.

Finally, we repeat this process for each pair of video frames (first frame from video 1 and second frame from video number 2). By applying the algorithm to each frame pair, we can create a target video where the foreground object from the first image is seamlessly combined with the background from the second image. This approach allows for the generation of visually appealing videos that incorporate specific foreground objects into different backgrounds.

\subsection{Image Segmentation Models}

Our next step involves utilizing image segmentation models to apply our algorithm to video frames and subsequently evaluate the model's performance.

\subsection{DETR (End-to-End Object Detection) model with ResNet-50 backbone}

The facebook/detr-resnet-50-panoptic model is an implementation of the DEtection TRansformer (DETR) for end-to-end object detection. It utilizes a ResNet-50 backbone for feature extraction. The model was trained on the COCO 2017 panoptic dataset, which consists of 118k annotated images \cite{res}.

The model can be used for panoptic segmentation, which involves predicting COCO classes, bounding boxes, and masks for objects in an image. It is designed for general object detection tasks and has been trained on a large-scale dataset. However, like any model, it has certain limitations, and its performance may vary depending on the specific use case \cite{res}.

We used this model to extract the target(s) and remove the background from our input image. you can see some samples in the following figures.

\begin{figure}[H]
\centering
\includegraphics[width=1\linewidth]{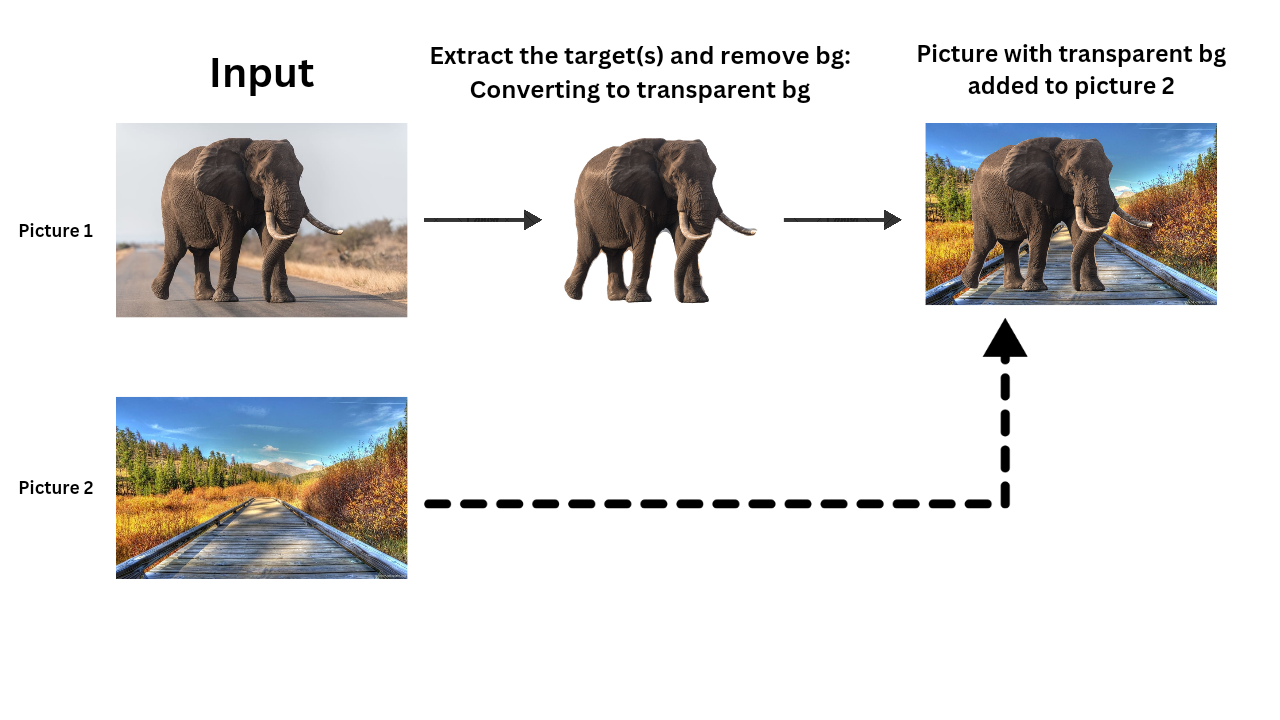}
\caption{\label{fig:pic1}Using detr-resnet-50-panoptic to extract the target and remove the background. We then added the picture with transparent background to the second input image.}
\end{figure}

\begin{figure}[H]
\centering
\includegraphics[width=1\linewidth]{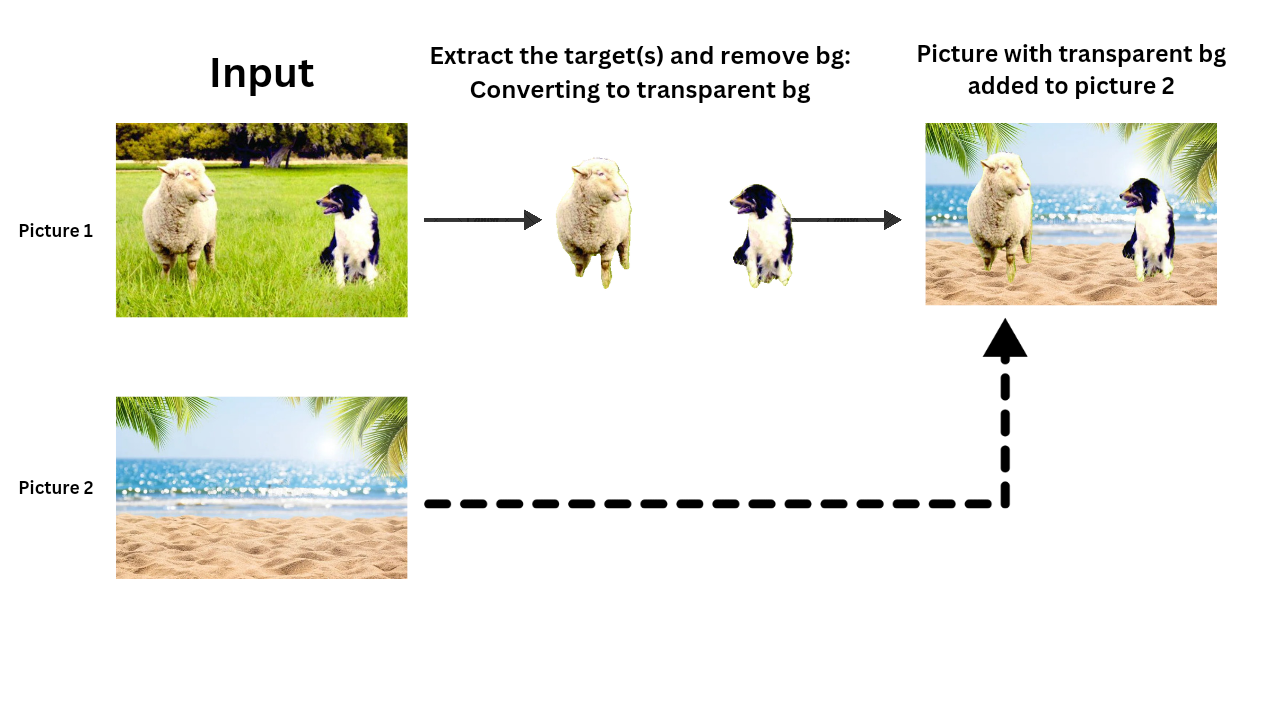}
\caption{\label{fig:pic2}The same thing as figure \ref{fig:pic1} one but this time applying the detr-resnet-50-panoptic model to extract multi-task from the target input image.}
\end{figure}

\subsubsection{Applying Image Segmentation to Videos}

In the context of video processing, the process of applying image segmentation involves converting video frames into images, applying the segmentation process to each image, and then reassembling the processed images back into a video. This approach treats each frame as an independent image and applies the segmentation and overlay logic to each frame. This process ensures that the complexity of the video is reduced, enabling further processing or analysis of each frame.

In the case of our program, the segmentation process involves extracting certain labels from the first video and converting their backgrounds to transparent. The resulting images are then overlaid onto corresponding frames from the second video. The labels selected for the first frame are used for subsequent frames, and if a label is not present in a particular frame, it is simply skipped for that frame. If a label reappears in a later frame, it will be included again as it remains in the list of selected labels.

\begin{figure}[H]
	\centering
	\begin{subfigure}{\linewidth}
		\includegraphics[width=\linewidth]{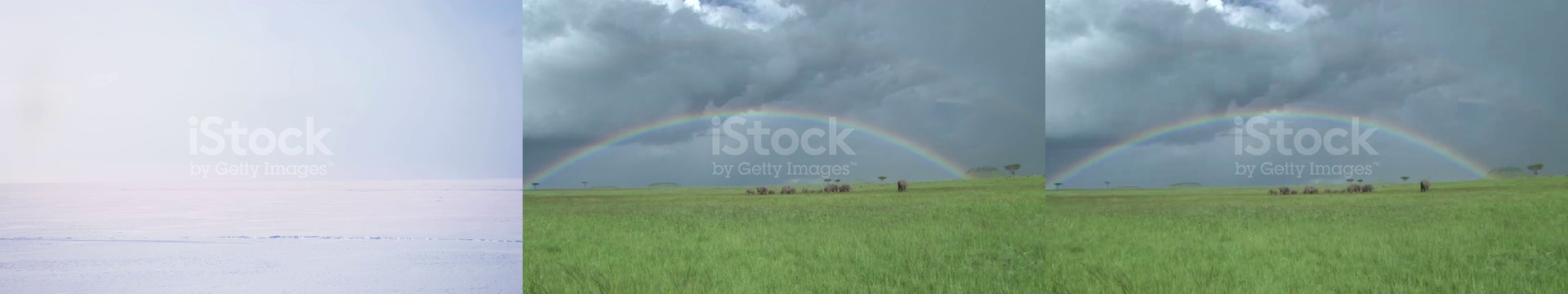}
	\end{subfigure} \\
	\begin{subfigure}{\linewidth}
		\includegraphics[width=\linewidth]{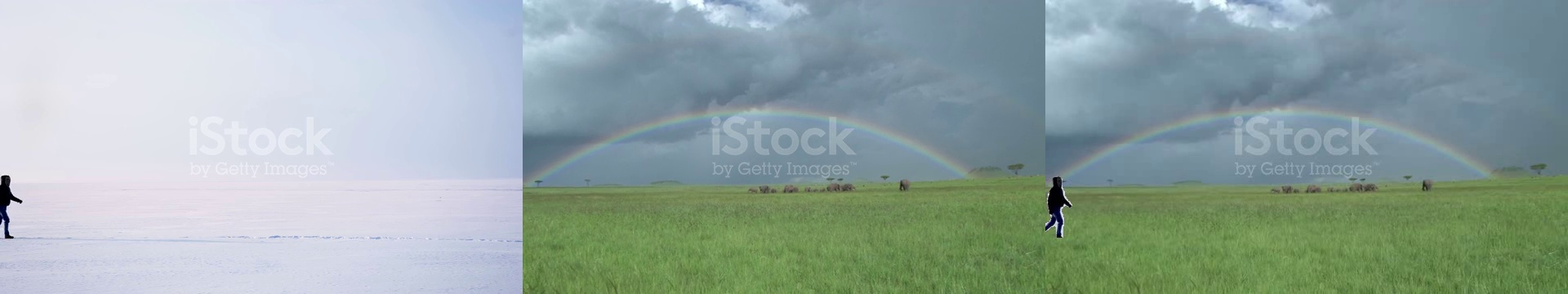}
	\end{subfigure}\\
 	\begin{subfigure}{\linewidth}
		\includegraphics[width=\linewidth]{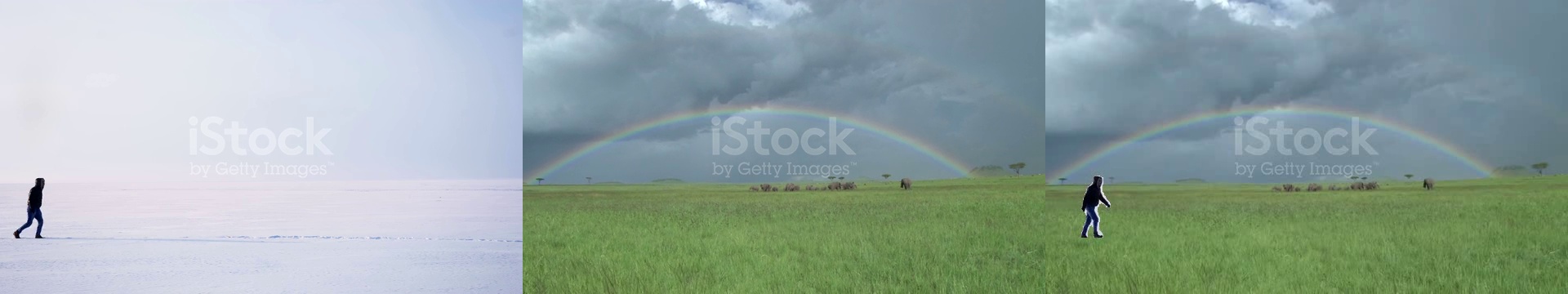}
	\end{subfigure}\\
	\begin{subfigure}{\linewidth}
		\includegraphics[width=\linewidth]{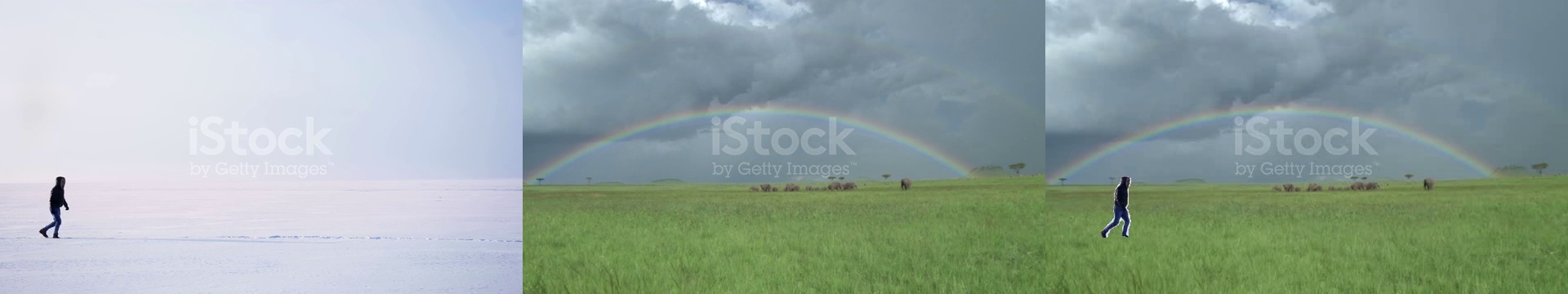}
	\end{subfigure}\\
	\begin{subfigure}{\linewidth}
	    \includegraphics[width=\linewidth]{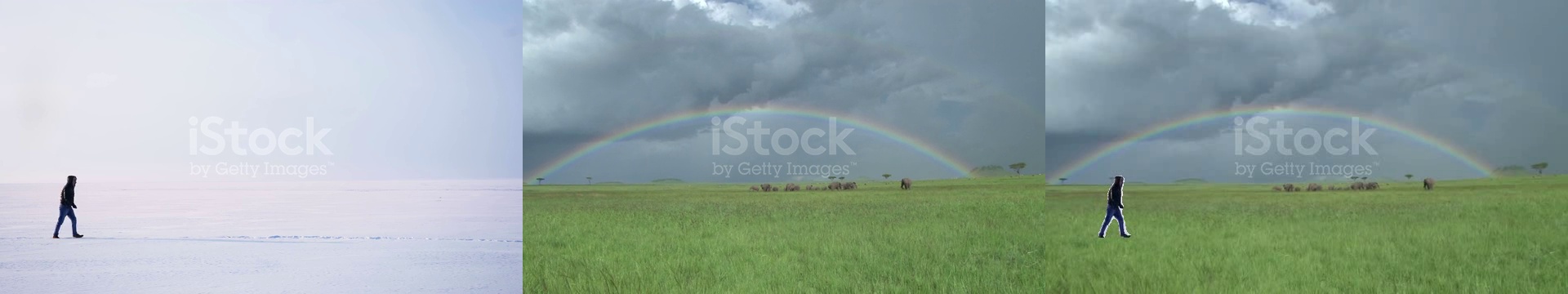}
	\end{subfigure}
	\caption{Applying image segmentation to video frames.}
	\label{fig:subfigures}
\end{figure}

It's important to note that the number of frames in the first video should match the number of frames in the second video. This ensures that each frame from the first video has a corresponding frame in the second video to serve as the background. However we handled this problem and user can use any input video for this task. The program also handles any exceptions that might occur due to mismatched frame sizes or out-of-range indices, preventing crashes due to unexpected input.

This approach to video processing leverages the power of image segmentation to create a unique visual experience. By allowing users to select specific labels for segmentation and overlay, the program offers a level of customization that enhances the user's control over the final output.

\subsubsection{Samples}

To provide a visual understanding of the process and results of the program, here are some sample inputs and outputs: \url{https://drive.google.com/drive/folders/1hnLLOXk25gv031NZaPkUbPyRTenwyPnN?usp=drive_link}.
These samples showcase the transformation of video frames using the image segmentation and overlay techniques explained earlier.

\subsubsection{Evaluating DETR Model with ResNet-50 Backbone: A Comparative Analysis with Faster R-CNN}

This model demonstrates impressive performance on the COCO 2017 validation dataset, boasting a box AP (average precision) score of 38.8, a segmentation AP score of 31.1, and a PQ (panoptic quality) score of 43.4 \cite{end}.

\begin{figure}[H]
\centering
\includegraphics[width=1\linewidth]{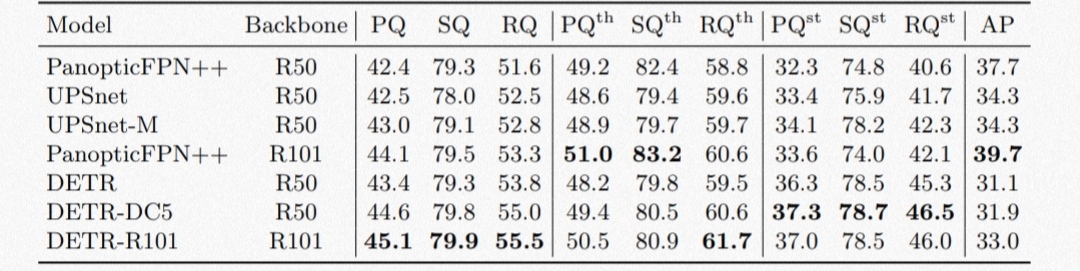}
\caption{\label{fig:pic22}Comparison with the state-of-the-art methods UPSNet and Panoptic FPN on the COCO val dataset they retrained PanopticFPN with the same data-augmentation as DETR, on a $18x$ schedule for fair comparison. UPSNet uses the $1x$ schedule, UPSNet-M is the version with multiscale test-time augmentations \cite{end}.}
\end{figure}

For the evaluation process, we can utilize the insights and methodologies presented in the research paper titled "End-to-End Object Detection with Transformers" authored by Nicolas Carion et al \cite{end}.

In the evaluation part of the paper, the authors thoroughly assess the performance of their proposed method, DETR, for object detection. They compare DETR with an optimized Faster R-CNN baseline on the COCO (Common Objects in Context) dataset.

The evaluation begins by describing the dataset used, which includes the COCO 2017 detection and panoptic segmentation datasets. These datasets consist of 118k training images and 5k validation images, each annotated with bounding boxes and panoptic segmentation. The average number of instances per image is 7, with a maximum of 63 instances in a single image in the training set.

To evaluate the performance of DETR, the authors utilize quantitative metrics, primarily the Average Precision (AP) and Intersection over Union (IoU). The AP is calculated as the integral metric over multiple thresholds, and the IoU measures the overlap between predicted bounding boxes and ground truth boxes.

The authors compare the performance of DETR with Faster R-CNN in terms of AP for bounding box detection. They report the validation AP at the last training epoch for comparison with Faster R-CNN. The results of the evaluation are presented in a Table, which compares DETR with state-of-the-art methods such as UPSNet and Panoptic FPN. The figure \ref{fig:pic22} shows the table of the Panoptic Quality (PQ), Semantic Quality (SQ), and Recognition Quality (RQ) scores for different models and backbones.

\begin{figure}[H]
\centering
\includegraphics[width=0.5\linewidth]{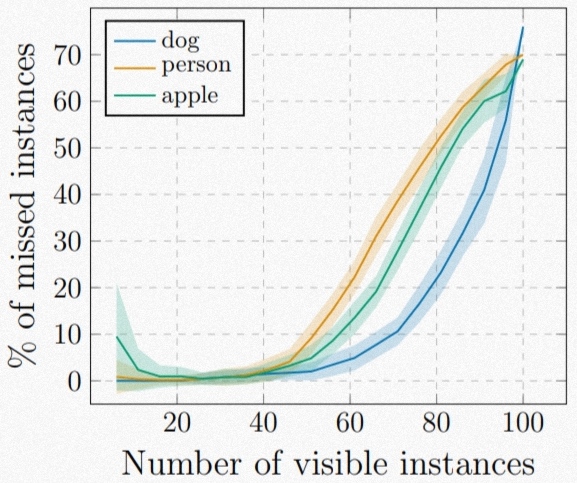}
\caption{\label{fig:pic33}Analysis of the number of instances of various classes missed by DETR de-pending on how many are present in the image. We report the mean and the standard deviation. As the number of instances gets close to 100, DETR starts saturating and misses more and more objects \cite{end}.}
\end{figure}

Additionally, the authors conduct an ablation study to analyze the impact of different components of the architecture and loss on the performance. They provide insights and qualitative results to gain a deeper understanding of DETR's strengths and weaknesses. The evaluation includes an analysis of the number of instances missed by DETR based on the number of visible instances in the image. Figure \ref{fig:pic33} illustrates this analysis, showing the percentage of missed instances for different classes as the number of instances increases.

The evaluation also involves analyzing the behavior of DETR when approaching the limit of its query slots. The authors create synthetic images with a grid of instances and evaluate the model's ability to detect them. The results show that DETR starts saturating and misses more instances as the number of instances approaches the limit.

\subsection{Utilization of the SegFormer B0 Model Fine-tuned on ADE20K}
As mentioned earlier in our research, we have chosen to utilize a state-of-the-art model for image segmentation tasks. In the previous subsection, we used the DETR model with ResNet-50 backbone. Now, we are exploring the performance of the SegFormer B0 model fine-tuned on the ADE20K dataset. This model, developed by NVIDIA, is hosted on the Hugging Face model hub \cite{nvidia}.

The SegFormer B0 is part of the SegFormer series of models, which are designed for segmentation tasks. These models are based on the Transformer architecture, which has shown significant performance in various domains, including natural language processing and computer vision. The Transformer architecture enables these models to capture long-range dependencies in data, which is beneficial in the context of image segmentation.

The specific model we used has been fine-tuned on the ADE20K dataset. The ADE20K is a widely used dataset for semantic segmentation and scene parsing tasks. It consists of over 20,000 images covering a wide variety of scenes and object categories. Fine-tuning the SegFormer B0 model on this dataset allows it to perform effectively across a broad range of segmentation tasks.

In our implementation, we integrated this model into our pipeline using the Hugging Face Transformers library. This library provides a straightforward interface to download and use pre-trained models, simplifying the integration of advanced machine learning capabilities into our project.

Now, we will evaluate the performance of the model on a single picture using the same elephant sample. We will assess its metrics to determine its effectiveness.

\begin{figure}[H]
\centering
\includegraphics[width=1\linewidth]{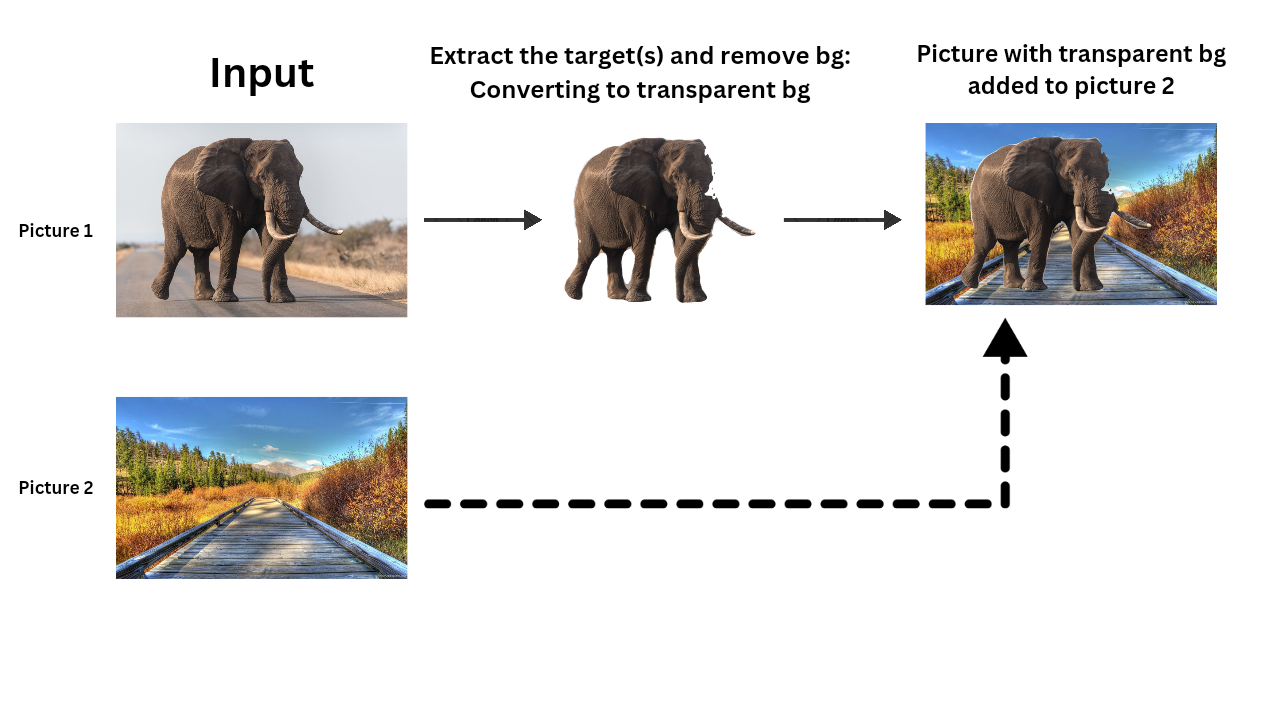}
\caption{\label{fig:pic3333}Same sample as figure \ref{fig:pic1} but this time using the SegFormer B0 Model Fine-tuned on ADE20K.}
\end{figure}

As illustrated in Figure \ref{fig:pic3333}, the performance of the SegFormer B0 Model, fine-tuned on the ADE20k dataset, does not quite match up to the output generated by the DETR model integrated with a ResNet-50 backbone (refer to Figure \ref{fig:pic1}). Despite this, the application of the ResNet-50 model to video frame data yielded superior results. The usage of SegFormer on our video samples can be further examined at the following link: 
\href{https://drive.google.com/drive/folders/1ajEYD2dIED7vjnvgkw0-FHZVNLyC-Pbs?usp=drive_link}{Google Drive Folder}.

\subsubsection{Evaluation and Metrics in SegFormer B0 model}

In the research paper titled "SegFormer: Simple and Efficient Design for Semantic Segmentation with Transformers" by Enze Xie, Wenhai Wang, Zhiding Yu, Anima Anandkumar, Jose M. Alvarez, and Ping Luo, the authors discuss the evaluation and metrics used to assess the performance of their proposed SegFormer framework for semantic segmentation \cite{nvidia}.

In terms of evaluation, the authors compare their results with existing approaches on two benchmark datasets: ADE20K and Cityscapes. For ADE20K, SegFormer-B0 achieves a mean Intersection over Union (mIoU) of $37.4\%$ using only 3.8 million parameters and 8.4 billion FLOPs. This outperforms all other real-time counterparts in terms of parameters, FLOPs, and latency. 

\begin{figure}[H]
\centering
\includegraphics[width=0.4\linewidth]{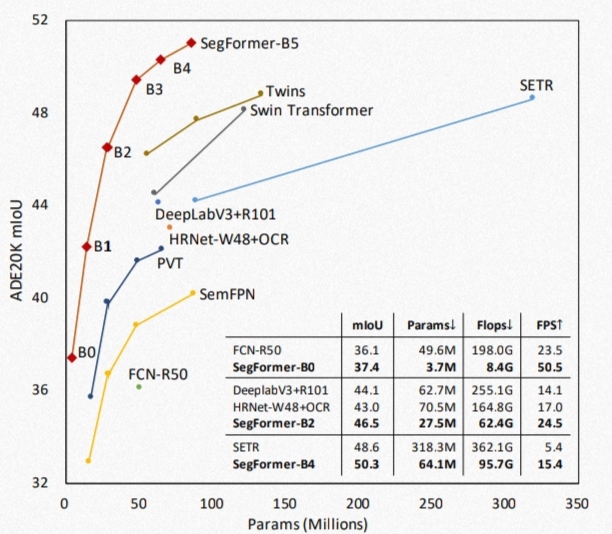}
\caption{\label{fig:pic44}Performance vs. model efficiency on ADE20K. SegFormer achieves a new state-of-the-art $51.0\%$ mIoU while being significantly more efficient than previous methods \cite{nvidia}.}
\end{figure}

On Cityscapes, SegFormer-B5 achieves state-of-the-art results with a mIoU of $51.8\%$, which is $1.6\%$ higher than the previous best method, SETR. The authors also provide comparisons with other state-of-the-art methods on these datasets, showcasing the superior performance of SegFormer.

In terms of metrics, the authors primarily use mIoU as the evaluation metric. mIoU measures the overlap between the predicted segmentation masks and the ground truth masks across different classes. It provides an overall assessment of the model's ability to accurately segment objects in the images. The authors report mIoU values for different models and compare them with other methods to demonstrate the effectiveness of SegFormer.

\begin{figure}[H]
\centering
\includegraphics[width=1\linewidth]{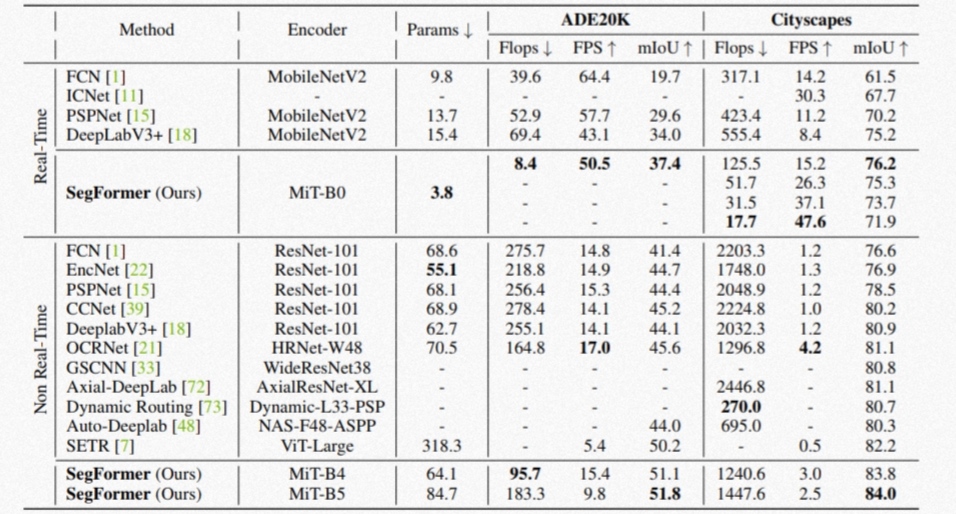}
\caption{\label{fig:pic55}Comparison to state of the art methods on ADE20K and Cityscapes. SegFormer has significant advantages on \#Params (M), \#Flops, \#Speed and \#Accuracy. Note that for SegFormer-B0 they scale the short side of image to {1024, 768, 640, 512} to get speed-accuracy tradeoffs. \cite{nvidia}.}
\end{figure}

Additionally, the authors discuss the influence of different decoders in their experiments. They find that the proposed MLP decoder in SegFormer has the least parameters compared to other decoders, such as UperNet decoder in Swin. This highlights the importance of the MLP decoder in achieving efficient segmentation.

\subsection{Segment Anything Model (SAM) by Hugging Face Transformers}

The Segment Anything Model (SAM) is a machine learning model developed by Facebook AI and implemented in the Hugging Face Transformers library. It is designed to predict segmentation masks for any object of interest given an input image, making it a versatile tool for image segmentation tasks \cite{sam}.

SAM uses an image processor, a prompt encoder, and a mask decoder, which are all configurable. The image processor pre-processes the image, the prompt encoder encodes the 2D points and bounding boxes, and the mask decoder generates the masks.

The image processor can perform various operations like resizing, rescaling, normalizing, padding, and converting the image to RGB.

The prompt encoder uses input 2D points and labels to encode the prompt. The labels indicate whether a point contains the object of interest, does not contain the object, corresponds to the background, or is a padding point that should be ignored.

The mask decoder generates the segmentation masks based on the encoded prompts and other parameters like image embeddings and attention similarity tensors.

The SAM model also supports optional features like returning attention tensors of all attention layers, hidden states of all layers, and a ModelOutput instead of a plain tuple.

Once the masks are generated, they can be post-processed by removing padding, upscaling to the original image size, and optionally binarizing them.

The SAM model can be used with any image and 2D point, and it can generate masks for multiple points in the same image. Although the original paper suggested that textual input should also be supported, this feature is not yet implemented in the official repository.

Now, we will evaluate the performance of the model on a single picture using the same elephant sample. We will assess its metrics to determine its effectiveness.

\begin{figure}[H]
\centering
\includegraphics[width=1\linewidth]{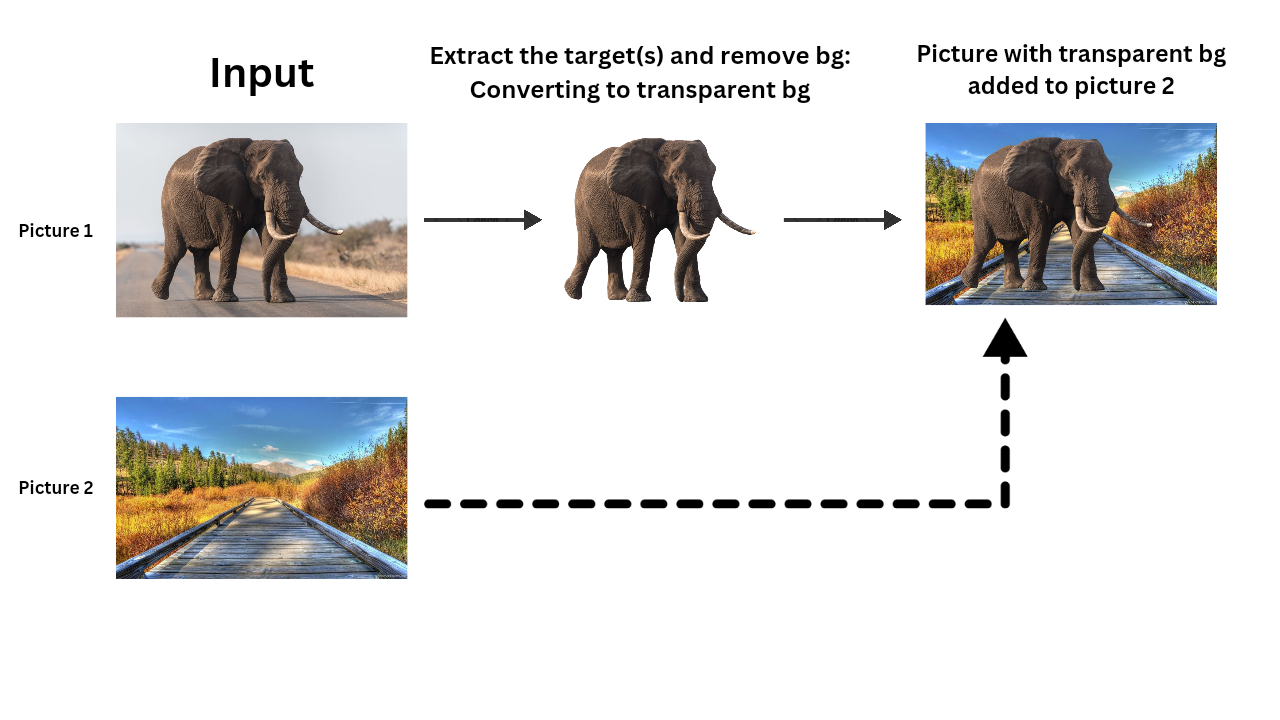}
\caption{\label{fig:pic33335}Same sample as figure \ref{fig:pic1} but this time using SAM.}
\end{figure}

The performance of the SAM model has proven to be excellent in our specific input sample, however, we have encountered a problem. For comparison, let's examine how Resnet-50 and SegFormer B0 models label elements in the same input picture.

Resnet-50 labels the elements as follows:
\begin{lstlisting}[style=DOS]
0: LABEL_187
1: road
2: LABEL_184
3: elephant
4: LABEL_193
\end{lstlisting}
The user is then prompted to enter the indices of the labels to retain, separated by a space.

Similarly, the SegFormer B0 model labels the elements in the input picture as:
\begin{lstlisting}[style=DOS]
0: sky
1: earth
2: rock
3: animal
\end{lstlisting}

Again, the user is prompted to enter the indices of the labels they wish to retain, separated by a space.

However, when we examine the labeling process of the SAM model, we encounter a problem. The model provides no named labels to select from. Instead, it provides a dictionary of values for each label, including the segmentation, area, bounding box, predicted Intersection over Union (IoU), point coordinates, stability score, and cropping box, as illustrated below:
\newpage
\begin{lstlisting}[style=DOS]
0: {
    'segmentation': array([[False, False, False, ..., False, False, False],
        [False, False, False, ..., False, False, False],
        [False, False, False, ..., False, False, False],
        ...
        [False, False, False, ..., False, False, False],
        [False, False, False, ..., False, False, False],
        [False, False, False, ..., False, False, False]]),
    'area': 939316,
    'bbox': [0, 5, 2047, 836],
    'predicted_iou': 1.0333043336868286, 
    'point_coords': [[32.0, 363.109375]], 
    'stability_score': 0.9882173538208008, 
    'crop_box': [0, 0, 2048, 1367]
}
\end{lstlisting}
This process continues for each label in the input image, leaving the user with no clear way to select the labels they wish to keep.

The SAM model generates binary masks for different segments in the image, but it doesn't provide the name of these segments. The model doesn't know what it's segmenting, it just tries to segment different objects or areas in the image. Therefore, it only returns a list of binary masks, where each mask corresponds to a different segment in the image \cite{why}.

This means that SAM doesn't provide labels for the segments, so you can't choose a segment by its name. If you want to select specific segments after running SAM, you would need to visually inspect the generated masks and decide which ones you want to keep based on their appearance, not their label. We intend to precisely execute this action.

\begin{figure}[H]
\centering
\includegraphics[width=0.9\linewidth]{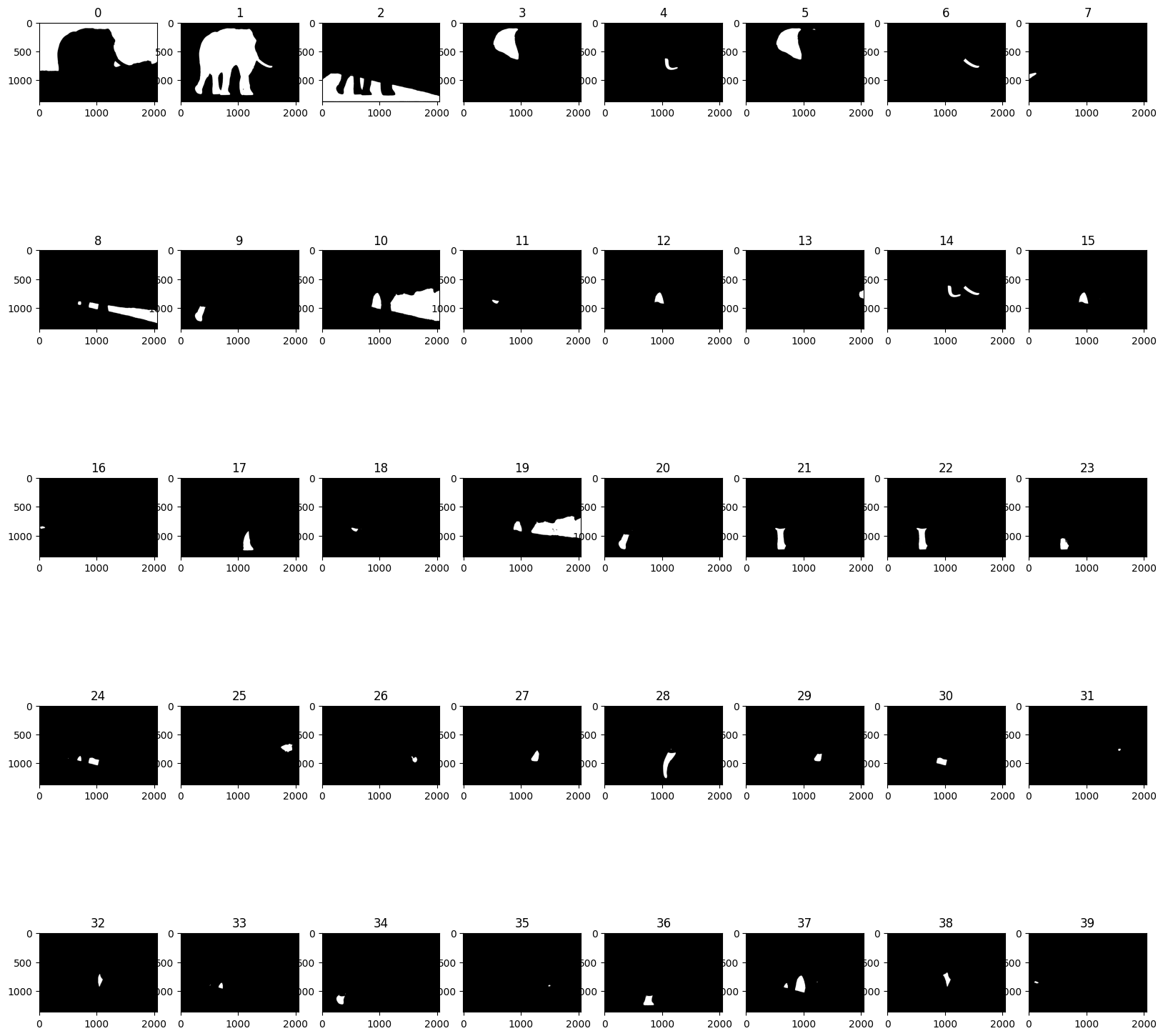}
\caption{\label{fig:mask1}Creating a grid of subplots and displaying each mask as a separate image in the grid. Also adding the index of each mask as the title of the corresponding subplot.}
\end{figure}

We created a grid of subplots and displayed each mask as a separate image in the grid. We also added the index of each mask as the title of the corresponding subplot and asked for user input after visualizing the picture (refer to Figure \ref{fig:mask1}).

\begin{figure}[H]
\centering
\includegraphics[width=0.9\linewidth]{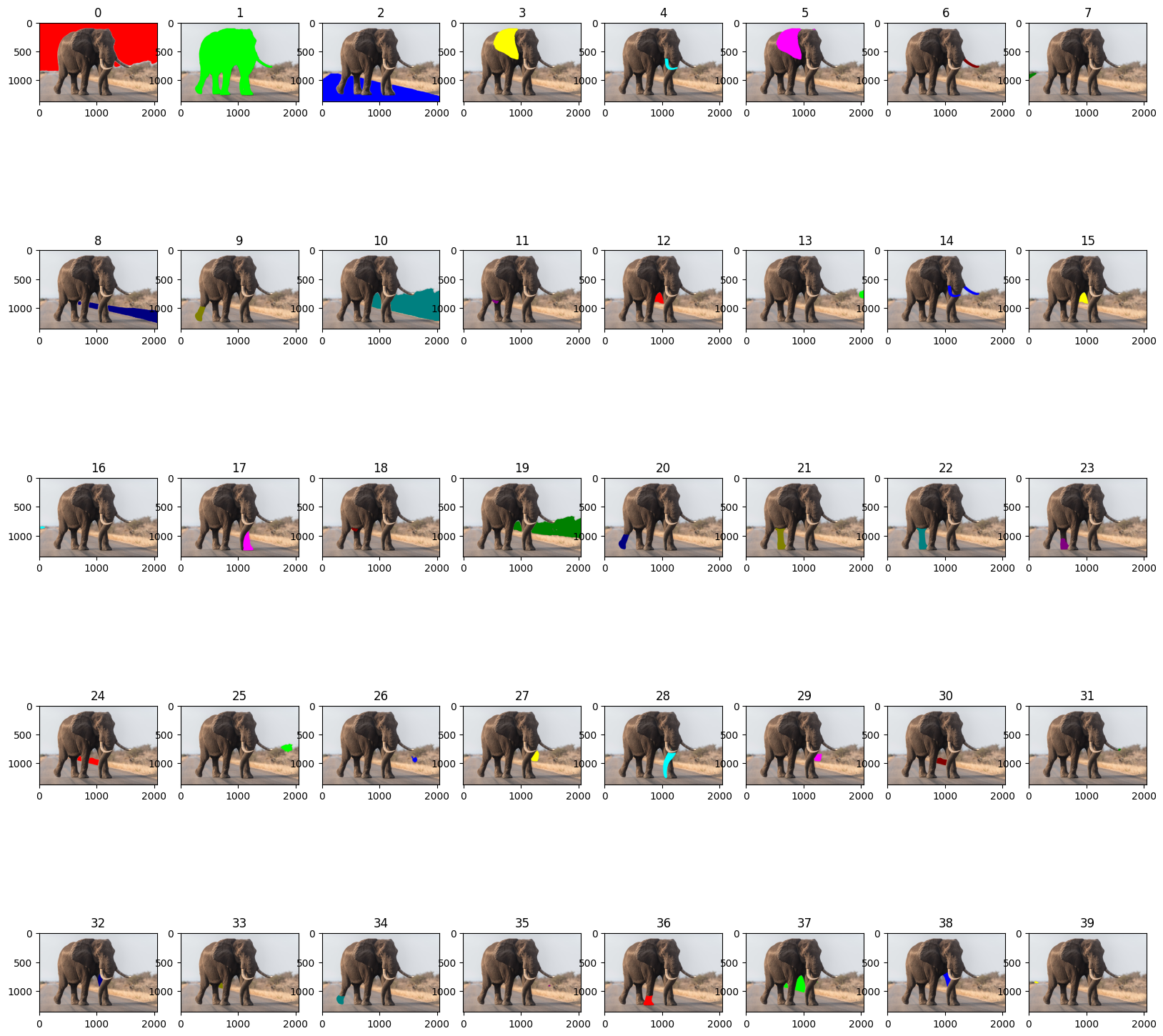}
\caption{\label{fig:mask2}Overlaying each mask onto the original image using a distinct color for each mask, and displaying the result as a separate image in the grid.}
\end{figure}

To enhance the visualization, we can overlay the masks on the original image. This will show the segmented parts in color while leaving the rest of the image as is (refer to Figure \ref{fig:mask2}). The primary challenge at hand is our uncertainty regarding whether the image we select to extract in the first frame will capture the same elements in subsequent frames, as we encountered a setback in this regard.

Another promising approach to implement SAM with video frames involves leveraging YOLOv8, a powerful model capable of predicting objects within our frames and generating labels accordingly. This method offers significant potential for improvement.

In my quest to apply the Segment-Anything-Model (SAM) to videos, I developed an algorithm. Although the results were not $100\%$ successful, the attempt was a solid starting point. In the subsequent section, I will provide a detailed technical report of the steps I took. You can find a sample of the work in this \href{https://drive.google.com/drive/folders/1rqCwFAIWh8ZK0TZUCnKjBSoVugCb3o4y}{Google Drive folder}.

Please keep an eye on this part of my GitHub repository, \href{https://github.com/Amirrezahmi/Video-Inpainting-and-Voice-Cloning/tree/main/video%20inpainting/applying%20IS%20to%20video}{applying IS to video}, as I will continue to work on this aspect to achieve the best possible results.

\subsection{Technical report}
%\subsubsection{Introduction}
In this report, we discuss a method for applying semantic segmentation to videos, using a combination of computer vision and machine learning techniques.

\subsubsection{Methodology}

Our method involves the use of a computer vision model known as SAM (Segment Anything Model) and a technique called Intersection over Union (IoU) for object tracking across video frames. SAM is a pre-trained deep learning model designed for semantic segmentation of images. It was chosen for its ability to generate multiple segmentation masks for a given image, providing a variety of segmentation possibilities.

For video segmentation, we applied the SAM model to each frame of the video, generating a list of segmentation masks for each frame. The user is then presented with a visual list of segmentation possibilities for the first frame and selects a mask. For subsequent frames, the program automatically selects the mask that most closely resembles the one chosen for the previous or first frame (based on the approach).

\subsubsection{Techniques and Metrics Used}

The key technique used in our method is the Intersection over Union (IoU) metric, a common evaluation metric in object detection and segmentation tasks. IoU measures the overlap between two binary masks, calculated as the area of intersection divided by the area of union of the two masks. We used the IoU metric to compare the segmentation masks of consecutive frames and select the mask that most closely matches the one from the previous or first frame (based on the approach).

Another important technique used in our method is the use of binary masks for image segmentation. A binary mask is a binary image that has the same spatial dimensions as the input image. In the mask, the pixels belonging to the object of interest are set to 1, while all other pixels are set to 0. This type of mask allows us to easily isolate the object of interest in the image.

\subsubsection{Results and Discussion}
The results showed that the method was able to successfully segment the videos and track the object of interest across frames.

However, there were some challenges encountered during the implementation:

\begin{enumerate}
  \item Applying SAM to video frames: The SAM model is designed for single image segmentation and does not inherently support video processing. We had to adapt the model to handle video frames by treating each frame as a separate image and applying SAM to each one.
  \item Selection of segmentation mask: Unlike traditional image segmentation models, SAM does not provide labels for segmentation. It generates several possible segmentations and the user needs to choose the one that best matches the object of interest. For video processing, we had to devise a method to automatically select the best mask for each frame. We came up with three approaches that we will talk about in the following.
  \item Dealing with moving objects: Since the object of interest can move or change shape over time in a video, the mask from the first frame might not be the best reference for the entire video. We had to adjust our mask selection method to use the mask of the previous frame as the reference, allowing the mask to adapt to changes in the object's position and shape over time.
  \item Ensuring object size consistency: We encountered an issue where the size of the object in the output video was smaller than in the original video. This was resolved by ensuring that the mask and the frame it's being pasted onto have the same size. We achieved this by resizing the mask to match the size of the frame before pasting it.
\end{enumerate}

Now we are going to talk about our approaches.
\subsubsection{Approach 1: Comparison with Previous Frame}

The first approach involves comparing each frame with the immediately preceding frame. The idea is to use the mask from the previous frame to select the most similar mask from the current frame's list of masks. The Intersection over Union (IoU) metric is used to measure the similarity between masks, and the mask with the highest IoU with the previous mask is chosen for the current frame.

However, this approach doesn't work well in certain scenarios. For instance, if the object of interest moves or changes significantly from one frame to the next, the mask from the previous frame might not be a good reference for the current frame. The IoU between the previous mask and the current masks could be low for all current masks, leading to poor selection of the mask for the current frame.

\subsubsection{Approach 2: Comparison with First Frame}

The second approach involves comparing each frame with the first frame. The mask from the first frame, chosen by the user, is used as a reference for all the subsequent frames. For each frame, the mask that most closely resembles the first mask is selected.

This approach works well in scenarios where the object of interest remains relatively consistent throughout the video. However, it may not work well when the object changes significantly over time, such as a person walking closer to the camera.

Moreover, if the scene changes completely (e.g., a cut to a different scene), comparing to the first frame would not make sense. The IoU scores would likely drop significantly because the object from the first frame is no longer present.

\subsubsection{Approach 3: Adaptive Reference Frame and User Input}

To address the limitations of the previous approaches, a third approach could be used that combines elements of both. This approach involves dynamically updating the reference frame based on certain conditions.

If the object of interest changes significantly over time, the reference frame could be updated periodically. For instance, a new reference frame could be chosen every N frames, or whenever the IoU with the current frame drops below a certain threshold. This would allow the reference to adapt to significant changes in the video.

If a completely new scene starts and the comparison with the first frame doesn't make sense anymore, the user could be prompted to select a new mask from the current frame. This would ensure that the chosen mask is always relevant to the current scene.

Despite these challenges, the method showed promising results in applying semantic segmentation to videos. With further tuning and optimization, it could potentially be used in a wide range of applications, from video editing to object tracking in surveillance videos. The precision of the model's segmentation results leads me to believe that further refinement could yield the desired results. This makes the model a potential alternative to the DETR (End-to-End Object Detection) model with a ResNet-50 backbone.

\subsubsection{\textcolor{red}{Why SAM is better than the previous models?}}

The object detection model we initially chose was the DETR (End-to-End Object Detection) model with a ResNet-50 backbone. However, upon further evaluation, we found that the SAM exhibited superior performance, and we have decided to proceed with SAM as our selected model.

To empirically demonstrate the superior capabilities of SAM, I applied both the DETR and SAM models to a set of low-quality videos. The DETR model's results were not as satisfactory, with the object detection appearing less realistic. In contrast, the SAM's results were outstanding, demonstrating its effectiveness in handling low-quality video data.

Here is a sample output from the fifth frame of the output video, processed by both these models:

\begin{figure}[H]
    \centering
    \begin{subfigure}[b]{0.48\linewidth}        %% or \columnwidth
        \centering
        \includegraphics[width=\linewidth]{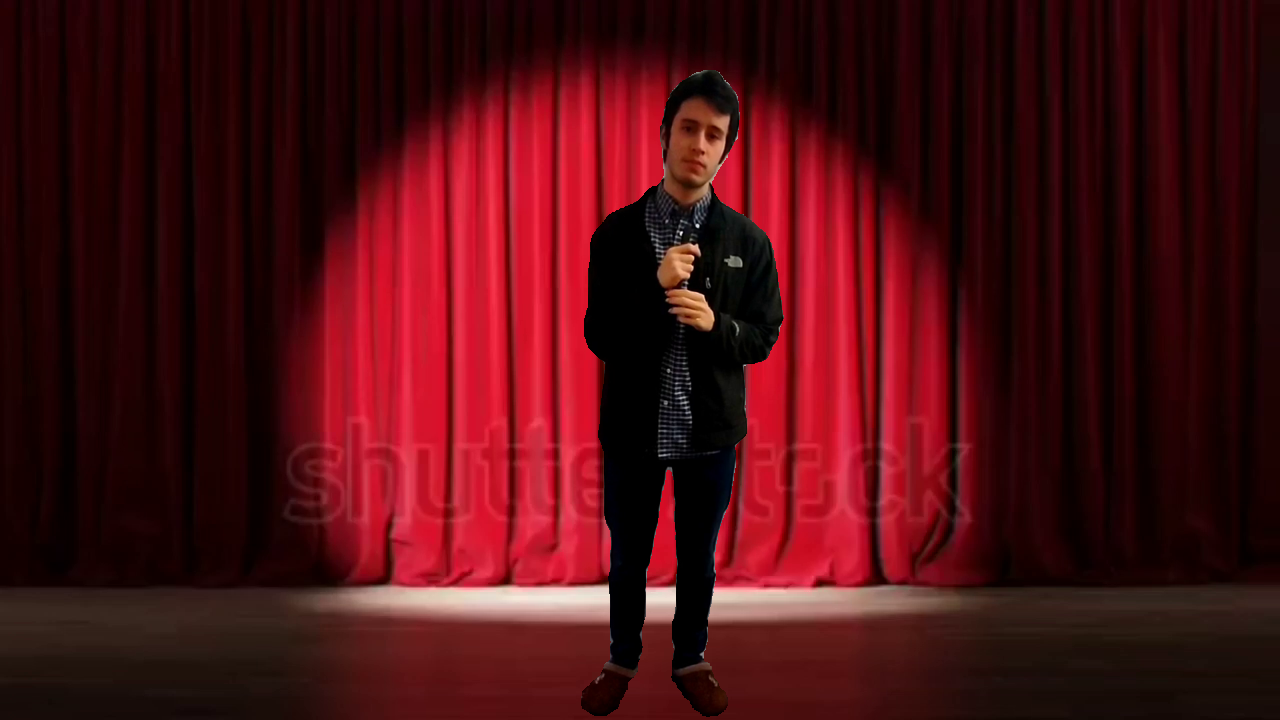}
        \caption{Output produced by the SAM.}
        \label{fig:A}
    \end{subfigure}
    \begin{subfigure}[b]{0.48\linewidth}        %% or \columnwidth
        \centering
        \includegraphics[width=\linewidth]{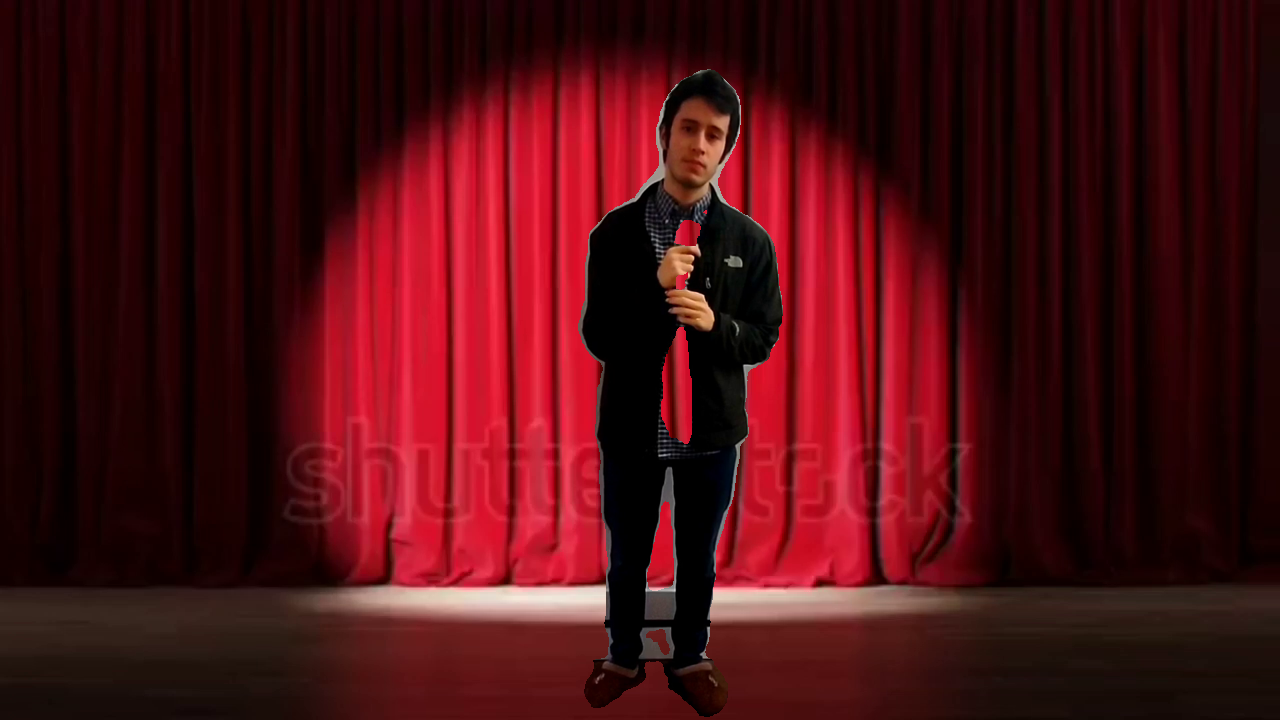}
        \caption{Output produced by the DETR.}
        \label{fig:B}
    \end{subfigure}
    \caption{Comparative Visualization of Image Segmentation Results: (a) SAM Model Output vs. (b) DETR Model Output}
    \label{fig:compare}
\end{figure}

As you can see in Figure \ref{fig:compare}, the output generated by SAM is much better than the output generated by DETR.

\subsubsection{Evaluation and Metrics in "Segment Anything": Comparisons and Results}

In the research paper titled "Segment Anything" by Alexander Kirillov, Eric Mintun, Nikhila Ravi, Hanzi Mao, Chloe Rolland, Laura Gustafson, Tete Xiao, Spencer Whitehead, Alexander C. Berg, Wan-Yen Lo, Piotr Doll´ar, and Ross Girshick, the authors discuss the evaluation and metrics used in their study. They introduce the Segment Anything Model (SAM) and evaluate its performance using various metrics and comparisons \cite{sam}.

\begin{figure}[H]
\centering
\includegraphics[width=1\linewidth]{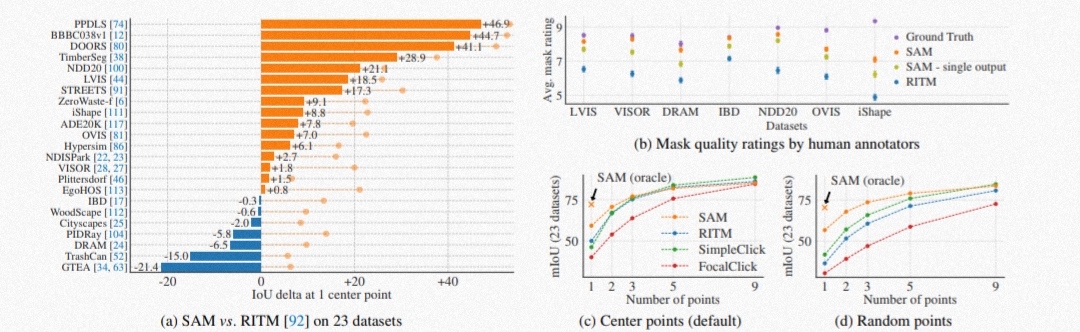}
\caption{\label{fig:mask2222}Point to mask evaluation on 23 datasets. (a) Mean IoU of SAM and the strongest single point segmenter, RITM. Due to ambiguity, a single mask may not match ground truth; circles show “oracle” results of the most relevant of SAM’s 3 predictions. (b) Per-dataset comparison of mask quality ratings by annotators from 1 (worst) to 10 (best). All methods use the ground truth mask center as the prompt. (c, d) mIoU with varying number of points. SAM significantly outperforms prior interactive segmenters with 1 point and is on par with more points. Low absolute mIoU at 1 point is the result of ambiguity \cite{sam}.}
\end{figure}

The evaluation in the paper focuses on image segmentation tasks and utilizes the SA-1B dataset, which contains over 1 billion masks on 11 million licensed and privacy-respecting images. The authors compare SAM to several baselines, including RITM, a strong interactive segmenter. They evaluate the models based on metrics such as mean Intersection over Union (mIoU) and mask quality ratings.

The mIoU metric is used to measure the similarity between predicted masks and ground truth masks. The authors evaluate SAM's performance at different points, ranging from 1 to 9, where points represent the number of prompts used for segmentation. They compare SAM's mIoU scores to those of the baselines, particularly RITM, and analyze the performance across different datasets.

\begin{figure}[H]
\centering
\includegraphics[width=0.7\linewidth]{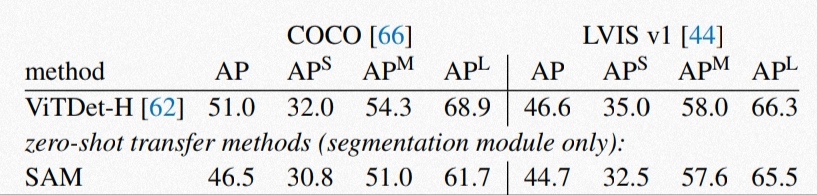}
\caption{\label{fig:t1}Instance segmentation results. SAM is prompted with ViTDet boxes to do zero-shot segmentation. The fully-supervised ViTDet outperforms SAM, but the gap shrinks on the higher-quality LVIS masks. Interestingly, SAM out-performs ViTDet according to human ratings (see Fig. \ref{fig:t2}) \cite{sam}.}
\end{figure}

\begin{figure}[H]
\centering
\includegraphics[width=0.7\linewidth]{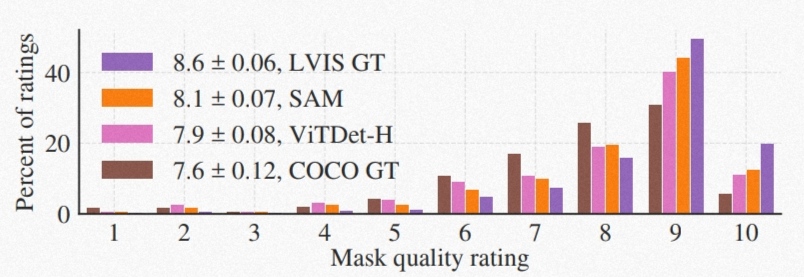}
\caption{\label{fig:t2}Mask quality rating distribution from their human study for ViTDet and SAM, both applied to LVIS ground truth boxes. We also report LVIS and COCO ground truth quality. The legend shows rating means and $95\%$ confi-dence intervals. Despite its lower AP (Fig. \ref{fig:t1}), SAM has higher ratings than ViTDet, suggesting that ViTDet exploits biases in the COCO and LVIS training data \cite{sam}.}
\end{figure}

Additionally, the authors conduct a human study to evaluate mask quality. Professional annotators rate the quality of masks generated by SAM and the baselines on a scale from 1 to 10. The authors compare the mask quality ratings between SAM and the baselines, as well as with ground truth masks.

The results of the evaluation and metrics show that SAM performs well in various downstream tasks, including edge detection, object proposal generation, and instance segmentation. SAM achieves competitive mIoU scores compared to the baselines, particularly RITM. The human study also indicates that SAM produces visually higher quality masks compared to the baselines, even when traditional metrics like mIoU do not reveal this difference.

\subsection{Stable Diffusion Models}

Stable Diffusion (SD) is a diffusion model used for image inpainting, a process that fills in missing or corrupted areas of an image. This model operates on the latent representations of images and can be implemented in different ways \cite{B02, A01}.

\subsubsection{Stable Diffusion Inpainting Model}
Stable Diffusion Inpainting Model uses a fine-tuned SD model for inpainting tasks. It substitutes the area not under the mask with the original image latents plus the amount of noise needed for this step. The area under the mask is left untouched. This way, SD only changes the masked area and knows about the area that needs to stay the same \cite{A01}.

\subsubsection{Diffusion-Based Inpainting Methods}

Diffusion-based inpainting methods propagate information from the surrounding regions of the image into the missing or corrupted areas. This approach is relatively simple to implement and computationally efficient compared to other techniques. However, it's important to ensure stability during the diffusion process to avoid artifacts such as ringing or blurring, which can degrade the inpainting quality \cite{B02}.

We will now delve into a comprehensive discussion of our algorithm while conducting an analysis of various pretrained stable diffusion models that have proven to be valuable in our video inpainting task.

\subsection{stable-diffusion-inpainting}

Our first model leads to a model from RunwayML called stable-diffusion-inpainting which is hosted on HuggingFace \cite{C03}. Our task focused on video inpainting, but we decided to utilize models designed for image inpainting, primarily because videos can be broken down into individual frames which are essentially images. Our goal was to feed two videos into the system: the first video featuring a subject (for example, a person), and the second video showcasing a scenery. The target video would then transport the subject from the first video into the background of the second video.

\begin{figure}[H]
\centering
\includegraphics[width=0.6\linewidth]{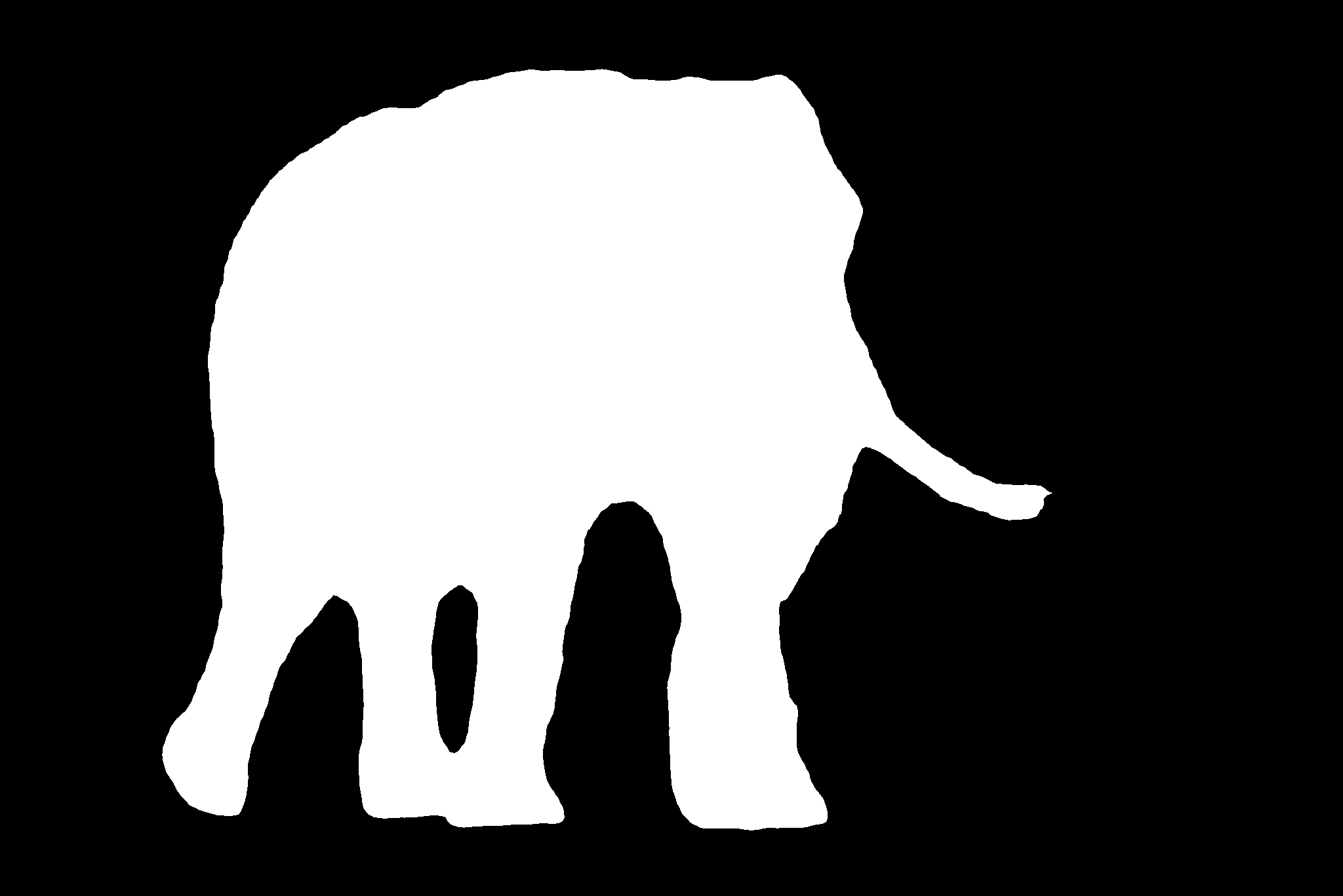}
\caption{\label{fig:ele1}Creating a binary mask of our target (with white pixels) using DETR model with ResNet-50 backbone.}
\end{figure}

The algorithm we used for this purpose was Stable Diffusion Inpainting. The process involved providing an image and a mask of the background of the first image. In the context of image inpainting, a mask is a binary image that indicates which parts of the original image should be inpainted. The mask is typically the same size as the input image, with areas to be inpainted marked as 1 (or True), and the remaining areas marked as 0 (or False) \cite{ee}. To automate this process, we can employ the models discussed in the image segmentation subsection, such as the SAM or the DETR model with a ResNet-50 backbone. This can be demonstrated using an example where a mask is created for a sample image of an elephant, as depicted in Figure \ref{fig:ele1}.

In Figure \ref{fig:ele1}, we have generated a binary mask for our elephant. This process allows us to isolate the elephant from the input image. The pixels representing the elephant are white, indicating a value of 1 or True, while the remaining pixels are black, signifying a value of 0 or False.

\begin{figure}[H]
\centering
\includegraphics[width=0.6\linewidth]{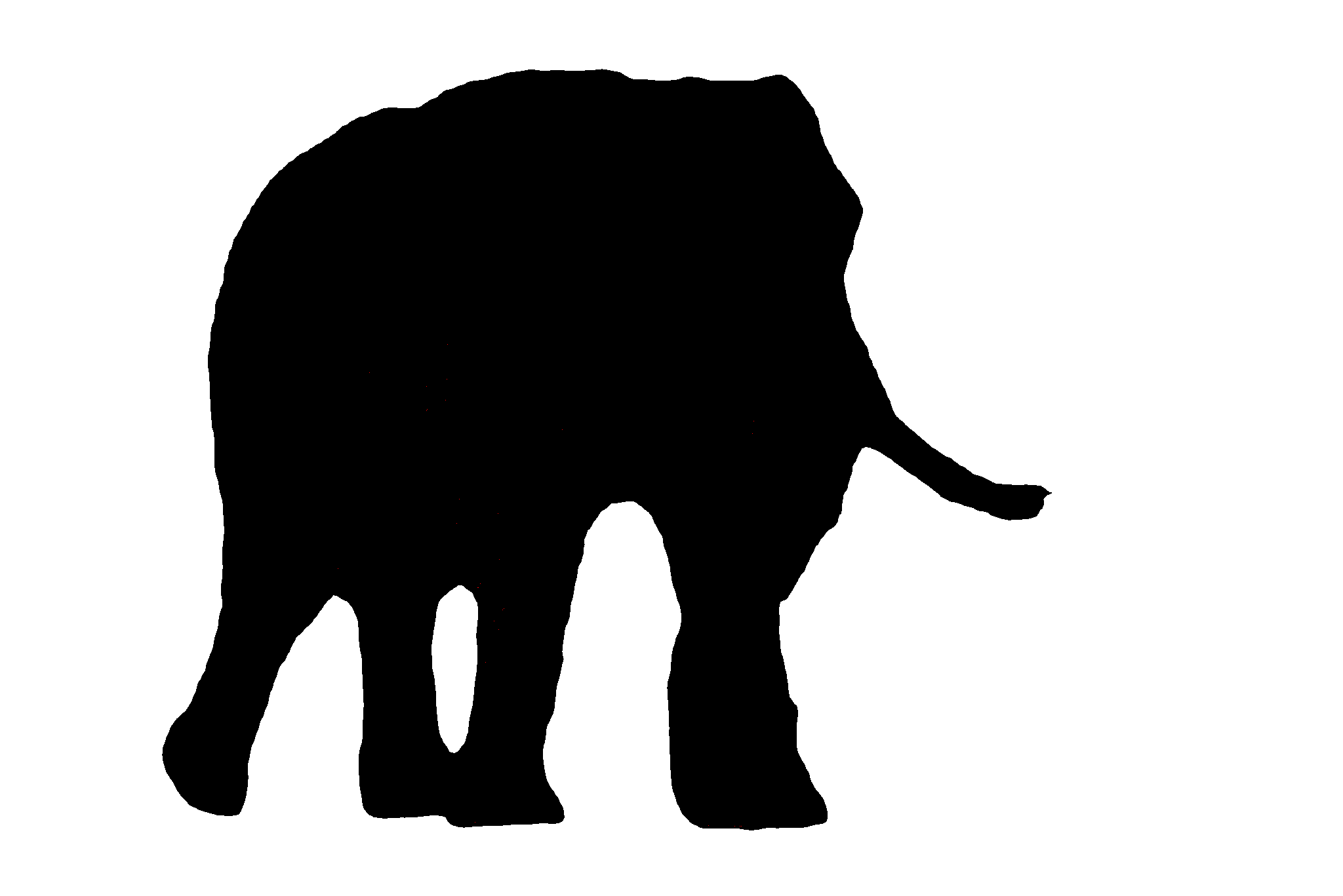}
\caption{\label{fig:ele2}Inverting the pixel values to make the background our new target.}
\end{figure}

Despite this, our objective is not to extract the target but rather its background. We aim to eliminate the background of our target object from the image. For instance, in our sample elephant image, after creating a binary mask and extracting the elephant (represented by a white image), we can invert the pixel values to make the background our new target. This means, pixels with a value of 1 become 0 and pixels with a value of 0 become 1. This operation is illustrated in Figure \ref{fig:ele2}.

By selecting the appropriate input prompt, such as 'change the background to transparent background', we could potentially eliminate the background, leaving only the main subject.

For the task of background removal, we initially attempted to use specific input prompts such as 'change the background to transparent background', with the intention of eliminating the background and retaining only the main subject. However, when we applied this approach using stable diffusion, it did not yield the desired results. 

\begin{figure}[H]
\centering
\includegraphics[width=1\linewidth]{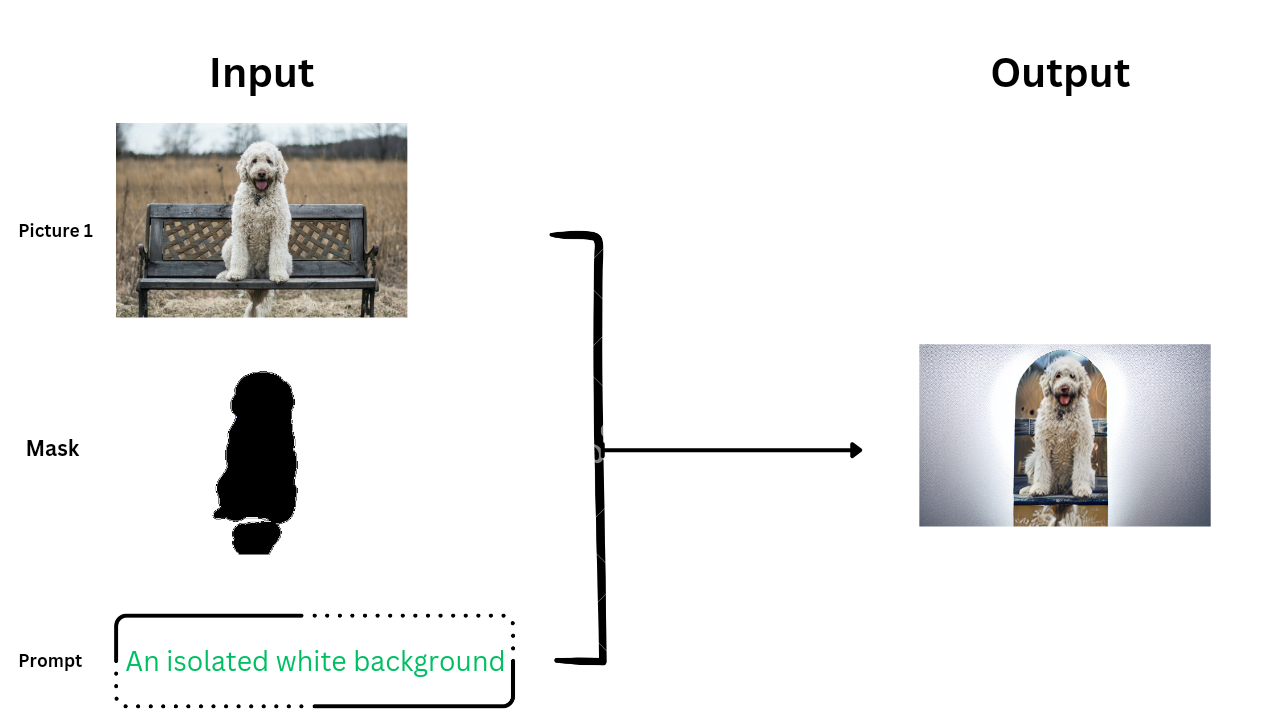}
\caption{\label{fig:sd1}Changing the background by the mask and input prompt using stable diffusion inpainting.}
\end{figure}

A subsequent attempt was made using an alternative prompt, 'completely white background', to isolate the target on a white background. Unfortunately, this too resulted in an output that was not completely white, falling short of our expectations (refer to Figure \ref{fig:sd1}). 

The official Colab notebook for Stable Diffusion inpainting is accessible \href{https://colab.research.google.com/github/huggingface/notebooks/blob/main/diffusers/in_painting_with_stable_diffusion_using_diffusers.ipynb}{here}. For your convenience, I have also uploaded a copy of this notebook to my GitHub repository.

This Colab notebook provides detailed steps on how to use Stable Diffusion for inpainting tasks.

The notebook includes the necessary code snippets and explanations to guide you through the process. It covers crucial aspects such as loading the model, defining the inpainting mask, setting the parameters, and running the inpainting process.

The copy of this notebook in my GitHub repository serves as a backup and provides an additional access point for developers who might face access issues with the main link.

if you're running the notebook in a different environment (I assume you're running it on colab), you may need to install the required libraries manually.

Therefore, we concluded that the stable diffusion inpainting method was unsuitable for this particular task. Despite this setback, we continued our exploration of other methods in the following subsection, titled 'Automatic1111'.

\subsection{Automatic1111}

The The Stable Diffusion WebUI, also referred to as AUTOMATIC1111 or A1111, is a Graphical User Interface (GUI) specifically designed for sophisticated image generation tasks. This GUI is particularly utilized in conjunction with Stable Diffusion, a technology that employs Artificial Intelligence for image generation. Our primary objective in this study was to demonstrate the process of background removal from images.

A comprehensive tutorial on the subject can be found in a YouTube video available at \href{https://youtu.be/Ki_ZcF_u23I?si=CW-Htp5JvDUHLk0X}{this link}. The tutorial elucidates the method of background removal from images through the utilization of a specific browser extension. The user is required to initially install this extension and subsequently apply it to the Stable Diffusion WebUI.

Upon successful installation of the extension, the user has the ability to generate an image and selectively remove the background by accessing the "remove background" option within the "send to extras" menu. Furthermore, the extension allows the user to select the preferred mode of background removal, such as human segmentation or U2 net.

\begin{figure}[H]
\centering
\includegraphics[width=1\linewidth]{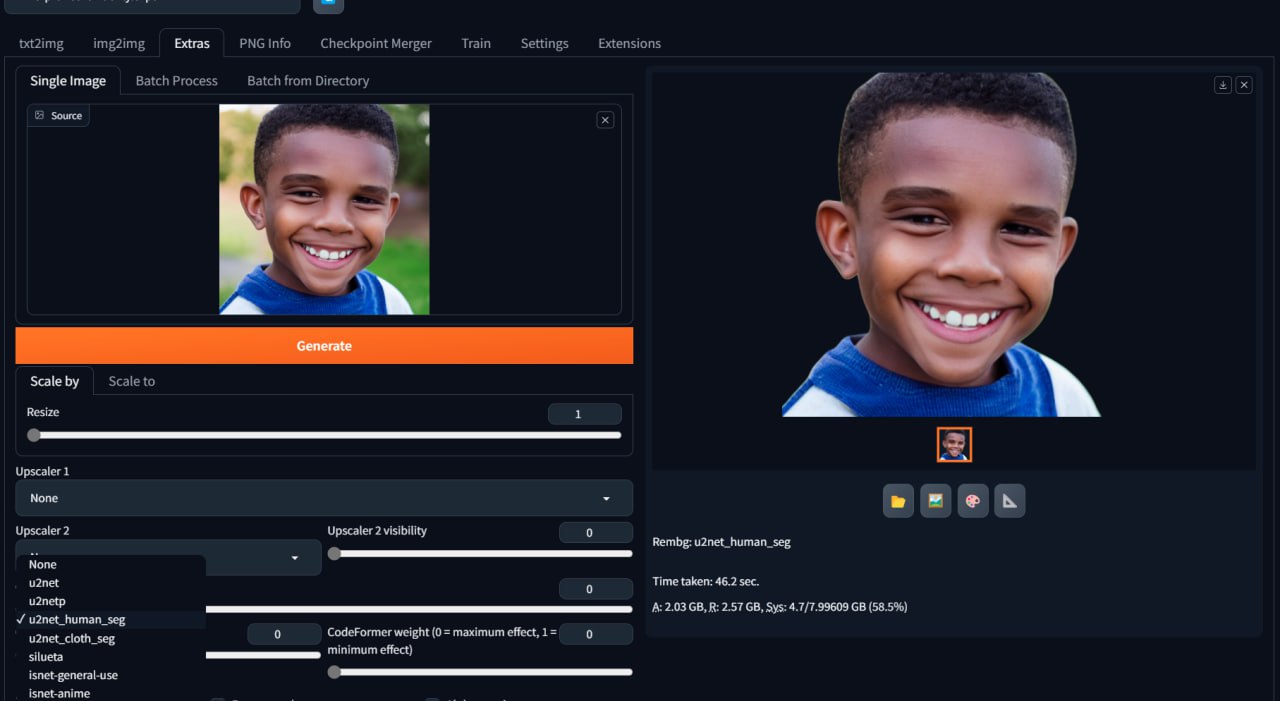}
\caption{\label{fig:a1111}Generating a picture of a boy smiling by our text prompt and then using the extension to remove the background in automatic1111 WebUI.}
\end{figure}

The tutorial provides a practical demonstration of the process, showing the removal of the background from an image of a bald man smiling against a forest backdrop. We did the same thing for an image of a boy smiling (refer to Figure \ref{fig:a1111}).

Utilizing a consistent methodology, we archived our video frames into a dedicated directory. This process was initiated by selecting the 'Batch from directory' option, which facilitated the application of the same procedure to every individual frame. Subsequently, we assembled a new video using the newly processed frames. The completed video can be accessed via the following link:

\url{https://drive.google.com/drive/folders/1fnmNeb2hoXePOosZS91GQ3KoFOXxiZb2}

However, we can use this extension directly without using a1111, as they have provided in their \href{https://github.com/danielgatis/rembg}{GitHub repository}. However, if you have seen the sample video, the quality is not better than the previous models.

\subsection{GitHub Repository - Access All Codes}

Our \href{https://github.com/Amirrezahmi/Video-Inpainting-and-Voice-Cloning}{GitHub repository} contains all the relevant source code for reference.

\subsection{Implementing a GPU-Accelerated Program on Linux Using Docker}

To run the program we implemented for DETR model with ResNet-50 back-bone on Linux, we decided to dockerize the program so anyone who wanted to run it can simply pull the image. We didn't dockerize the implementation of SAM to videos because it's not done yet but if I get the best results I will push it to my docker hub.

This guide will provide detailed steps on how to run a GPU-accelerated program on Ubuntu 22.04 using Docker, with a specific focus on the program located in the Docker Hub repository \textbf{amirrezahashemi/dockervi:v1m}. The program is Dockerized for ease of deployment and portability and utilizes GPU acceleration for improved performance.

\subsubsection{Prerequisites}

\begin{figure}[H]
\centering
\includegraphics[width=0.8\linewidth]{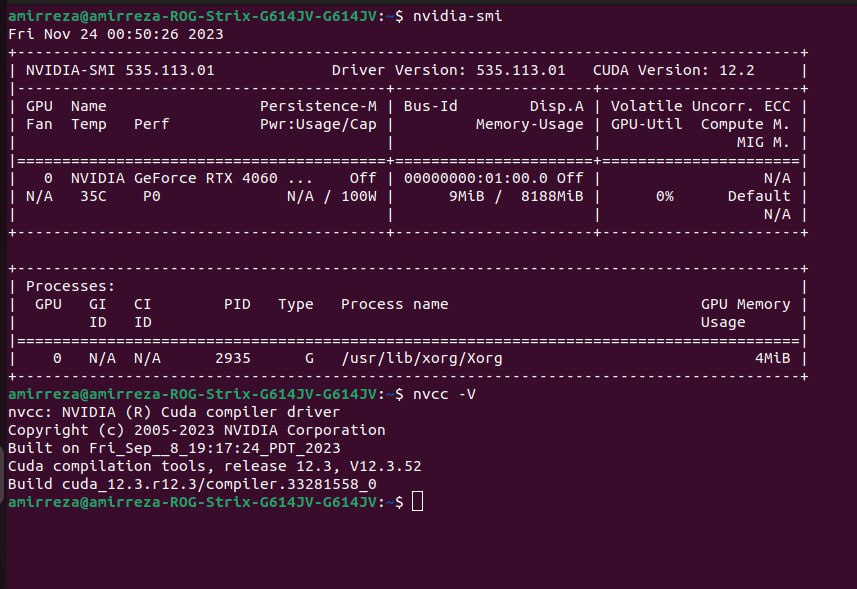}
\caption{\label{fig:piccx22}Ensuring that we have installed the CUDA toolkit and the Nvidia driver.}
\end{figure}

Before you can run the program, ensure that you have installed the CUDA toolkit and the Nvidia driver on your system, as the program requires GPU acceleration. You can verify the installation by executing the following commands in your terminal:

\begin{lstlisting}[style=DOS]
nvidia-smi
nvcc -V
\end{lstlisting}

These commands should return information about your GPU and CUDA version respectively (refer to Figure \ref{fig:piccx22}). If they do not return the expected output, it indicates that these components are not installed on your system. You can refer to the official Nvidia and Docker documentation, as well as this \href{https://github.com/HFarkhari/Docker/blob/main/Docker_presentation.pdf}{PDF} guide for installation instructions.

\subsubsection{Pulling the Docker Image}

Once you have confirmed that your system is properly configured, you can proceed to pull the Docker image of the program from Docker Hub. If you do not have a Docker Hub account, you can create one \href{https://hub.docker.com/}{here}. After creating an account, log in to Docker Hub by running the following command in the terminal:

\begin{lstlisting}[style=DOS]
docker login -u your_username
\end{lstlisting}

With successful login, you can now pull the Docker image:

\begin{lstlisting}[style=DOS]
docker pull amirrezahashemi/dockervi:v1m
\end{lstlisting}

\subsubsection{Running the Docker Image}

After you have successfully pulled the image, you can run it with the following command:

\begin{lstlisting}[style=DOS]
docker run -v realpath/Desktop/videos:/opt/app/videos --gpus all -it dockervi:v1m python vi.py
\end{lstlisting}

Alternatively, you can pull and run the image in a single command:

\begin{lstlisting}[style=DOS]
docker run --gpu all -it --rm -v $(realpath ~/Desktop/videos):/opt/app/videos amirrezahashemi/dockervi:v1m python vi.py
\end{lstlisting}

These commands mount the local \textbf{videos} directory to the \textbf{/opt/app/videos} directory inside the Docker container and allocate all available GPUs to the container. The \textbf{-it} flag allows interactive processes (like a shell), and the \textbf{--rm} flag automatically removes the container when it exits.

\subsubsection{Accessing the Output}

The output video will be saved in the \textbf{videos} directory. If the output video is locked, you can unlock it by navigating to the \textbf{videos} directory on your Desktop and running the following command:

\begin{lstlisting}[style=DOS]
chmod 777 -R videos
\end{lstlisting}

If you opened a new terminal window, to navigate to the \textbf{videos} directory run the following command:

\begin{lstlisting}[style=DOS]
cd videos 
\end{lstlisting}

Then run the following command:

\begin{lstlisting}[style=DOS]
chmod 777 -R videos
\end{lstlisting}

This command changes the permissions of the \textbf{videos} directory, allowing all users to read, write, and execute files in the directory.

If you're unable to open the video on Ubuntu 22.04, you may need to remove the \textbf{gstreamer1.0-vaapi} package by running:

\begin{lstlisting}[style=DOS]
sudo apt remove gstreamer1.0-vaapi
\end{lstlisting}

This solution is suggested in this \href{https://askubuntu.com/questions/1406254/after-installing-ubuntu-22-04-the-default-video-player-is-unable-to-play-any-vi#comment2512000_1406276}{Ask Ubuntu post}. If you're still experiencing issues, consider trying a different video player.

\section{Voice Conversion}

The goal of Voice Conversion is to modify an audio input, known as the source voice, to sound like a different, target voice. This process can be applied to a variety of audio types, ranging from simple speech samples to complex singing samples. The task typically involves the use of machine learning models, which are trained to learn the characteristics of the target voice and apply these characteristics to the source voice.

\subsection{so-vits-svc-fork Model}

\subsubsection{Introduction}

The first model we are considering is the so-vits-svc-fork model, a fork of the so-vits-svc model developed by the SVC Develop Team. Both of these models can be found on GitHub.

The \href{https://github.com/voicepaw/so-vits-svc-fork}{so-vits-svc-fork model}, found on the voicepaw repository, is an enhanced version of the \href{https://github.com/svc-develop-team/so-vits-svc}{original model}, featuring real-time support and a greatly improved interface.

The original so-vits-svc model, hosted on the svc-develop-team repository, focuses on Singing Voice Conversion (SVC) rather than Text-to-Speech (TTS).

\subsubsection{Methodology}

The so-vits-svc-fork model uses a SoftVC content encoder to extract speech features from the source audio. These feature vectors are directly fed into VITS without the need for conversion to a text-based intermediate representation. As a result, the pitch and intonations of the original audio are preserved.

The model also employs a vocoder, specifically the NSF HiFiGAN, to solve the problem of sound interruption. The vocoder is a key component of the model, used to generate the final audio output that resembles the target voice.

\subsubsection{Installation and Usage}

For installation and usage please watch \href{https://youtu.be/xgvT7UnUTng?si=qTmFftcb8Z1gKC0b}{this video}.

\noindent
To train and use the So-Vits-SVC on your local system please watch \href{https://youtu.be/mnszz9W0kEg?si=pai9MaMPwnTdm3bS}{this video}.

\subsubsection{Evaluation and Metrics}

The performance of the so-vits-svc-fork model can be evaluated using a variety of metrics. One common metric for voice conversion tasks is the Mel Cepstral Distortion (MCD), which measures the difference between the Mel cepstral coefficients of the original and converted voices. Lower MCD values indicate better performance, as they signify a smaller difference between the original and converted voices.

Another metric that might be used is the Gross Pitch Error (GPE), which measures the difference in pitch between the original and converted voices. Like the MCD, lower GPE values indicate better performance.

However, it's worth noting that these objective metrics may not always align with subjective human perception of voice similarity and quality. Therefore, subjective listening tests are often conducted as well, where human listeners rate the similarity and quality of the converted voices.

Unfortunately, specific evaluation results or metrics for the so-vits-svc-fork model were not provided in the GitHub repositories.

Furthermore, the so-vits-svc-fork model has been used in a variety of applications, as seen on Hugging Face's model hub. These applications provide a practical perspective on the model's performance and usage.

For example, the model 1asbgdh/sovits4.0-volemb-vec768 on Hugging Face's model hub has been used for Audio-to-Audio tasks. Similarly, the model zhaijifu67/so-vits-4.0-collect has also been used for Audio-to-Audio tasks.

Read \href{https://thelearness.com/how-to-clone-any-voice-with-ai-with-so-vits-svc-fork/}{this} article for more details.

\subsection{Samples}

You can listen to some samples from the following link:

\url{https://drive.google.com/drive/folders/1WI3D0DMJ20VqoHZYsa6rztD1cMVqyfvg?usp=sharing}

\subsection{Codes}
You can access the code for this part from the \href{https://github.com/voicepaw/so-vits-svc-fork}{official} GitHub repository or you can use it from the notebook I pushed in my \href{https://github.com/Amirrezahmi/Video-Inpainting-and-Voice-Cloning/tree/main/voice%20cloning}{GitHub repository} as I added some additional parts to the notebook which are essential for our program, like exporting the vocal sound to instrumental sound.

\subsection{Retrieval-based-Voice-Conversion-WebUI}

The \href{https://github.com/RVC-Project/Retrieval-based-Voice-Conversion-WebUI}{Retrieval-based-Voice-Conversion-WebUI} is a simple and easy-to-use voice conversion framework based on VITS. The project is developed in Python and it supports multiple languages including English, Chinese, Japanese, Korean, French, and Turkish.

\subsubsection{Features}

The project offers several key features:

\begin{enumerate}
    \item Top1 retrieval replaces the input source features with training set features to eliminate timbre leakage.
    \item It allows for fast training even on relatively poor graphics cards.
    \item It can produce good results with a small amount of training data (it is recommended to collect at least 10 minutes of low-noise voice data).
    \item It allows for model fusion to change the timbre (with the help of ckpt processing tab in ckpt-merge).
    \item It offers a simple and easy-to-use web interface.
    \item The UVR5 model can be called to quickly separate vocals and accompaniment.
    \item It uses the most advanced vocal pitch extraction algorithm InterSpeech2023-RMVPE to eliminate mute problems.
    \item It supports acceleration for A-card and I-card.
\end{enumerate}
\subsubsection{Environment Setup}
The project requires Python version greater than 3.8. Dependencies can be installed using pip or poetry. For Nvidia cards, AMD cards, Intel cards, and Mac users, different sets of dependencies are required, which can be installed using the respective requirements text files.

\subsubsection{Pretrained Models}

The project requires some pretrained models for inference and training. These models can be downloaded from Hugging Face space. The list of all required pretrained models and other file names can be found in the repository.

\subsubsection{Usage}
To start the WebUI, you can run \textbf{python infer-web.py} command. For Windows or macOS users, you can directly download and unzip RVC-beta.7z, and then run \textbf{go-web.bat or sh ./run.sh} to start the WebUI.

\subsubsection{AMD Graphics Card Rocm (Only for Linux)}
If you want to run RVC on Linux systems based on AMD's Rocm technology, you need to install the necessary drivers. You could use pacman to install the required drivers if you are using Arch Linux:

\begin{lstlisting}[style=DOS]
pacman -S rocm-hip-sdk rocm-opencl-sdk
\end{lstlisting}

 For some models of graphics cards, you may need to configure the following environment variables: \textbf{ROCM\_PATH} and \textbf{HSA\_OVERRIDE\_GFX\_VERSION}. Also, make sure your current user is in the \textbf{render} and \textbf{video} user groups:
\begin{lstlisting}[style=DOS]
export ROCM_PATH=/opt/rocm
export HSA_OVERRIDE_GFX_VERSION=10.3.0
sudo usermod -aG render $USERNAME
sudo usermod -aG video $USERNAME
\end{lstlisting}

Finally start the WebUI by running the \textbf{python infer-web.py} command.

\subsubsection{Metrics}
The repository does not provide explicit details about the metrics used for evaluating the performance of the voice conversion model. However, as we said common metrics used in voice conversion tasks include Mel Cepstral Distortion (MCD), Fundamental Frequency (F0) RMSE and correlation, voiced/unvoiced error rate, and modulation spectrum.

\subsubsection{\color{red}What are the differences between RVC and SO-VITS-SVC models?}
This is the question that you may ask which has already been answered \href{https://ai.stackexchange.com/questions/40927/what-are-the-differences-between-rvc-and-so-vits-svc-models}{here}.

The answer expressed that these two models architectures are nearly the same, except that in SoVITS you can select which content encoder to use (HuBERT or ContentVec). They took the architecture of SoftVC and combine it with the design of VITS. The RVC is a succession of SoVITS and it has some improvements.

\begin{figure}[H]
\centering
\includegraphics[width=0.8\linewidth]{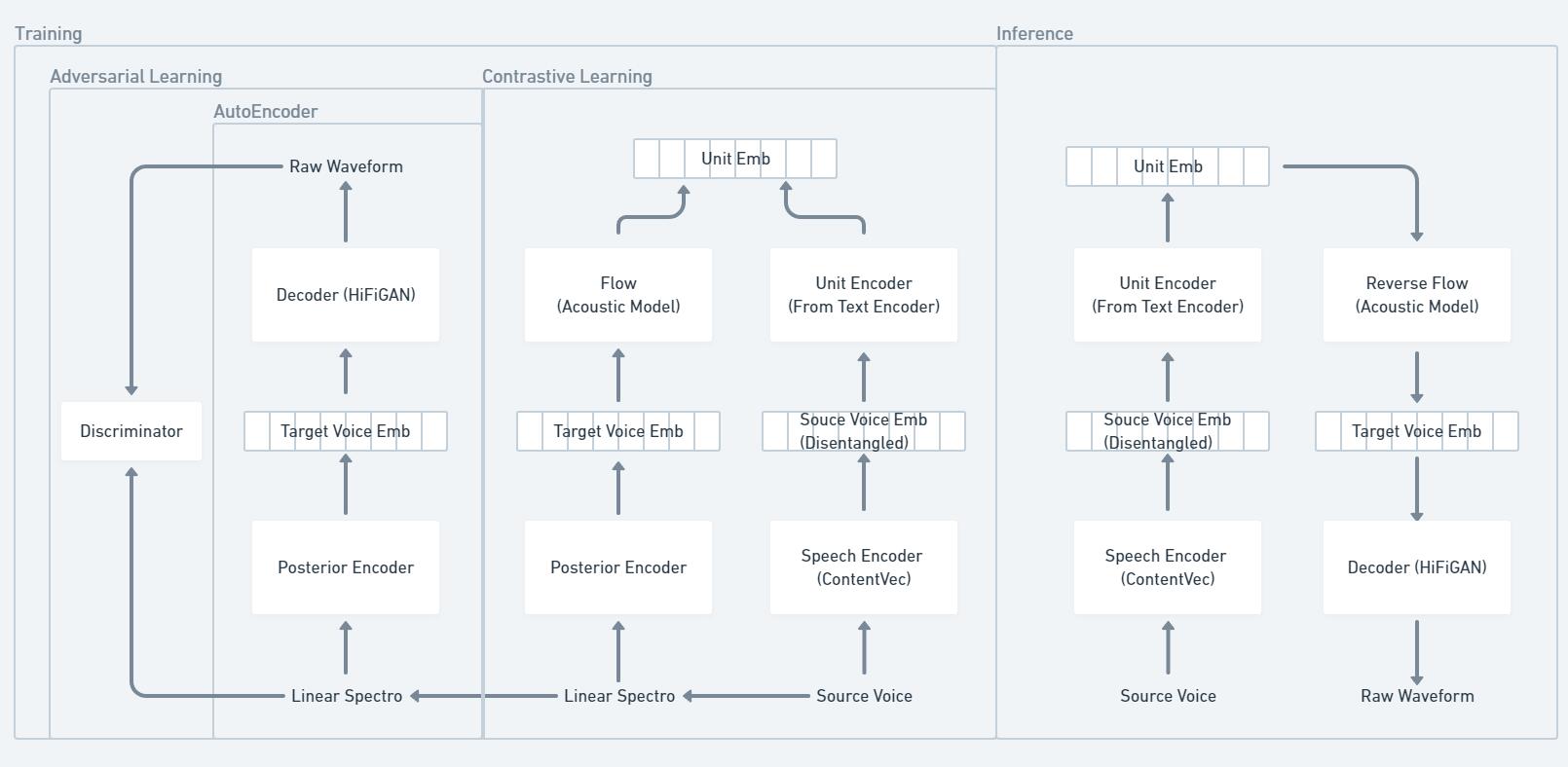}
\caption{\label{fig:Mmm}Illustrating the graph of the model of RVC.}
\end{figure}

Firstly, the RVC used ContentVec as the content encoder rather than HuBERT. ContentVec is an improved version of HuBERT, and it can ignore speaker information and only focus on content.

Secondly, the RVC used top1 retrieval to reduce tone leakage. It is just like the codebook used in VQ-VAE, mapping the unseen input into known input in the training dataset.

But according to the 4.1 version update in the SoVITS repo, they replaced HuBERT with ContentVec and also added feature retrieval functionality, so their performance should be the same now.
\subsection{Autovc}

The \href{https://github.com/auspicious3000/autovc}{AutoVC} repository provides a PyTorch implementation of the AutoVC framework, a many-to-many non-parallel voice conversion system. It's designed for zero-shot voice style transfer using only an autoencoder loss.

\subsubsection{Dependencies}
The project depends on the \textbf{wavenet\_vocoder} package, which can be installed using pip:

\begin{lstlisting}[style=DOS]
pip install wavenet_vocoder
\end{lstlisting}
\subsubsection{Usage}
The usage of the AutoVC project can be divided into three main steps:
\begin{enumerate}
    \item Convert Mel-Spectrograms: Download the pre-trained AutoVC model and run the \textbf{conversion.ipynb} in the same directory.
    \item Mel-Spectrograms to Waveform: Download the pre-trained WaveNet Vocoder model and run the \textbf{vocoder.ipynb} in the same directory.
    \item Train Model: The project includes a small set of training audio files in the wav folder, but it is recommended to prepare your own dataset for training. Generate spectrogram data from the wav files using \textbf{python make\_spect.py} and generate training metadata, including the GE2E speaker embedding, using \textbf{python make\_metadata.py}.
\end{enumerate}
\subsubsection{Troubleshooting}
If you encounter problems implementing AutoVC, consider using this fork: \href{https://github.com/KnurpsBram/AutoVC_WavenetVocoder_GriffinLim_experiments/tree/master}{\small AutoVC\_WavenetVocoder\_GriffinLim\_experiments}. This fork offers experiments on AutoVC and WaveNet vocoder, compared against the Griffin Lim spectrogram inversion algorithm. The experiments are broken down into steps to make them more understandable. The repository also provides a clear procedure for naming your experiment output files.

\subsection{yourtts}

The \href{https://github.com/Edresson/YourTTS}{YourTTS} repository provides an implementation of a multilingual approach to zero-shot multi-speaker Text-to-Speech (TTS) and voice conversion. This project builds upon the VITS model and introduces several modifications for zero-shot multi-speaker and multilingual training.

\subsubsection{Dependencies}

The project is implemented on the Coqui TTS repository, and all experiments were conducted there.

\subsubsection{Usage}
The usage of YourTTS can be divided into two main applications:
\begin{itemize}
    \item Text-to-Speech: You can use the released YourTTS model for TTS with the following command: 
\end{itemize}
\begin{lstlisting}[style=DOS]
tts  --text "This is an example!" --model_name tts_models/multilingual/multi-dataset/your_tts  --speaker_wav target_speaker_wav.wav --language_idx "en"
\end{lstlisting}
Here, "target\_speaker\_wav.wav" is an audio sample from the target speaker
\begin{itemize}
    \item Voice Conversion: You can use the released YourTTS model for voice conversion with the following command:
\end{itemize}
\begin{lstlisting}[style=DOS]
tts --model_name tts_models/multilingual/multi-dataset/your_tts  --speaker_wav target_speaker_wav.wav --reference_wav  target_content_wav.wav --language_idx "en"
\end{lstlisting}
Here, "target\_content\_wav.wav" is the reference wave file to convert into the voice of the "target\_speaker\_wav.wav" speaker.
\subsubsection{Metrics}
The project uses two key metrics for evaluating the performance of the YourTTS model, Mean Opinion Score (MOS) and Speaker Encoder Cosine Similarity (SECS).
\begin{enumerate}
    \item MOS: The MOS metric is commonly used for subjective quality assessment of telecommunication services. It provides a numerical indication of the perceived quality from the user's perspective. You can download the MOS samples from the link provided in the \href{https://github.com/Edresson/YourTTS/tree/main/metrics/MOS}{MOS directory}. You can recalculate the MOS with their respective confidence intervals for the English language using the following command:
\begin{lstlisting}[style=DOS]
python3 compute_similarity_MOS.py --csv_path EN/naturalness-MOS.csv
\end{lstlisting}
    \item SECS: This measures the cosine similarity between speaker embeddings. To reproduce the SECS results, you can run the notebooks present in the notebooks/ directory or use the Colab notebooks provided in the \href{SECS directory}{SECS directory}.
\end{enumerate}
You can find all the codes and details in their GitHub repository.

\subsubsection{Combination of Video Segmentation project and Voice Conversion project}
Upon the completion of this project, we have intriguingly integrated the results of the video segmentation and voice conversion components. The source code for this integration is available on \href{https://github.com/Amirrezahmi/Video-Inpainting-and-Voice-Cloning/tree/main/combination%20of%20vi%20%26%20vc}{GitHub}. Additionally, sample outputs from the project can be accessed on \href{https://drive.google.com/drive/u/0/mobile/folders/1WI3D0DMJ20VqoHZYsa6rztD1cMVqyfvg?usp=sharing}{Google Drive}.

The results from selected methodologies employed in our project can also be found on the aforementioned \href{https://drive.google.com/drive/u/0/mobile/folders/1WI3D0DMJ20VqoHZYsa6rztD1cMVqyfvg?usp=sharing}{Google Drive link}.

\section{Conclusion}

Throughout this study, we have explored and evaluated various models for image segmentation, image inpainting and voice conversion, aiming to create a unified solution named AutoVisual Fusion Suite. We utilized models such as SAM and the so-vits-svc-fork model for video segmentation and voice conversion. Additionally, we evaluated image inpainting using the stable Diffusion inpainting model. 
The application of SAM facilitated successful object tracking across frames, despite encountering certain challenges.
Further, our analysis of the so-vits-svc-fork model highlighted its potential for voice conversion. This comprehensive evaluation has laid the groundwork for future enhancements and applications, combined with the seamless integration of video segmentation and voice conversion methodologies.
The code, sample outputs, and results from this project are available for further exploration on GitHub and Google Drive. We look forward to the continued development and refinement of these models and techniques in future work.

\section{Acknowledgements}

I would like to express my deepest appreciation to Hamed Farkhari, whose invaluable guidance has significantly contributed to the success of my research project. His profound expertise in Linux and Docker technologies, along with his dedication to mentoring, have been instrumental in bringing my project to fruition.

Hamed's assistance in the Linux part of my project was critical. He guided me through the process of installing necessary tools such as the CUDA Toolkit and Nvidia drivers, which are essential for high-performance computing tasks on Linux. His step-by-step explanations and patient support helped me overcome numerous challenges, making the process much more manageable.

Moreover, Hamed's knowledge of Docker was equally beneficial. He demonstrated how to implement my project on Docker and push it to Docker Hub.

Beyond his technical expertise, Hamed also provided invaluable guidance in the research part of my project. His insights and suggestions greatly influenced the direction of my research, and his constant encouragement kept me motivated throughout the process.

Without Hamed's guidance and persistent help, this project would not have been possible. His contributions to my project were significant and for that, I am profoundly grateful.
\printbibliography

@misc{
    why,
    author = {Sovit Rath},
    title = {Segment Anything – A Foundation Model for Image Segmentation},
    year = {2023},
    url = {https://learnopencv.com/segment-anything/},
    note = {Accessed: 2023-10-09}
}

@misc{sam,
      title={Segment Anything}, 
      author={Alexander Kirillov and Eric Mintun and Nikhila Ravi and Hanzi Mao and Chloe Rolland and Laura Gustafson and Tete Xiao and Spencer Whitehead and Alexander C. Berg and Wan-Yen Lo and Piotr Dollár and Ross Girshick},
      year={2023},
      eprint={2304.02643},
      archivePrefix={arXiv},
      primaryClass={cs.CV}
}

@article{nvidia,
  author    = {Enze Xie and
               Wenhai Wang and
               Zhiding Yu and
               Anima Anandkumar and
               Jose M. Alvarez and
               Ping Luo},
  title     = {SegFormer: Simple and Efficient Design for Semantic Segmentation with
               Transformers},
  journal   = {CoRR},
  volume    = {abs/2105.15203},
  year      = {2021},
  url       = {https://arxiv.org/abs/2105.15203},
  eprinttype = {arXiv},
  eprint    = {2105.15203},
  bibsource = {dblp computer science bibliography, https://dblp.org}
}

@misc{end,
      title={End-to-End Object Detection with Transformers}, 
      author={Nicolas Carion and Francisco Massa and Gabriel Synnaeve and Nicolas Usunier and Alexander Kirillov and Sergey Zagoruyko},
      year={2020},
      eprint={2005.12872},
      archivePrefix={arXiv},
      primaryClass={cs.CV}
}

@article{res,
  author    = {Nicolas Carion and
               Francisco Massa and
               Gabriel Synnaeve and
               Nicolas Usunier and
               Alexander Kirillov and
               Sergey Zagoruyko},
  title     = {End-to-End Object Detection with Transformers},
  journal   = {CoRR},
  volume    = {abs/2005.12872},
  year      = {2020},
  url       = {https://arxiv.org/abs/2005.12872},
  archivePrefix = {arXiv},
  eprint    = {2005.12872},
  bibsource = {dblp computer science bibliography, https://dblp.org}
}

@misc{hg,
  author = {Hugging Face},
  title = {Image Segmentation Model},
  howpublished = "\url{https://huggingface.co/tasks/image-segmentation}"
}

@article{ee,
  title={Image inpainting: A review},
  author={Elharrouss, Omar and Almaadeed, Noor and Al-Maadeed, Somaya and Akbari, Younes},
  journal={Neural Processing Letters},
  volume={51},
  pages={2007--2028},
  year={2020},
  publisher={Springer}
}

@InProceedings{C03,
        author    = {Rombach, Robin and Blattmann, Andreas and Lorenz, Dominik and Esser, Patrick and Ommer, Bj\"orn},
        title     = {High-Resolution Image Synthesis With Latent Diffusion Models},
        booktitle = {Proceedings of the IEEE/CVF Conference on Computer Vision and Pattern Recognition (CVPR)},
        month     = {June},
        year      = {2022},
        pages     = {10684-10695}
}

@misc{A01,
  author = {Vadim Titko},
  year = {2023},
  title = {Let’s Understand Stable Diffusion Inpainting},
  howpublished="\url{https://medium.com/aibygroup/lets-understand-stable-diffusion-inpainting-fdd0b1c3a925}",
  journal={AIBY}
}

@misc{B02,
  author = {Daisie Team},
  year = {2023},
  title = {Understanding Stable Diffusion Models: A Comprehensive Guide
Understanding Stable Diffusion Models: A Comprehensive Guide},
  howpublished="\url{https://blog.daisie.com/understanding-stable-diffusion-models-a-comprehensive-guide/}"
}

@article{greenwade93,
    author  = "George D. Greenwade",
    title   = "The {C}omprehensive {T}ex {A}rchive {N}etwork ({CTAN})",
    year    = "1993",
    journal = "TUGBoat",
    volume  = "14",
    number  = "3",
    pages   = "342--351"
}

\end{document}